\documentclass{article}

\usepackage[preprint]{neurips_2025}
\usepackage[ruled,linesnumbered]{algorithm2e}
\usepackage[dvipsnames]{xcolor}
\usepackage{tcolorbox}
\usepackage{multirow}
\usepackage{colortbl}
\usepackage{wrapfig}
\usepackage{subfigure}
\usepackage{amsmath}

\usepackage[utf8]{inputenc} 
\usepackage[T1]{fontenc}    
\usepackage{hyperref}       
\usepackage{url}            
\usepackage{booktabs}       
\usepackage{amsfonts}       
\usepackage{nicefrac}       
\usepackage{microtype}      
\usepackage{xcolor}         

\title{FedGraM: Defending Against Untargeted Attacks in Federated Learning via Embedding Gram Matrix }

\author{%
  Di Wu\\
  School of Computer Science and Technology\\
  Xi'an Jiaotong University\\
  Xi'an 710071, China \\
  \texttt{ddiwu98@163.com} \\
  \And
  Qian Li\\
  School of Cyber Science and Engineering\\
  Xi'an Jiaotong University\\
  Xi'an 710071, China \\
  \texttt{qianlix@xjtu.edu.cn} \\
  \And
  Heng Yang\\
  School of Computer Science and Technology\\
  Xi'an Jiaotong University\\
  Xi'an 710071, China \\
  \texttt{yanghengjmpnow@gmail.com} \\
  \And
  Yong Han\\
  School of Computer Science and Technology\\
  Xi'an Jiaotong University\\
  Xi'an 710071, China \\
  \texttt{han\_yong@stu.xjtu.edu.cn} \\
  }

\begin{document}

\maketitle

\begin{abstract}
  Federated Learning (FL) enables geographically distributed clients to collaboratively train machine learning models by sharing only their local models, ensuring data privacy. However, FL is vulnerable to untargeted attacks that aim to degrade the global model's performance on the underlying data distribution. Existing defense mechanisms attempt to improve FL's resilience against such attacks, but their effectiveness is limited in practical FL environments due to data heterogeneity. On the contrary, we aim to detect and remove the attacks to mitigate their impact. Generalization contribution plays a crucial role in distinguishing untargeted attacks. Our observations indicate that, with limited data, the divergence between embeddings representing different classes provides a better measure of generalization than direct accuracy. In light of this, we propose a novel robust aggregation method, FedGraM, designed to defend against untargeted attacks in FL. The server maintains an auxiliary dataset containing one sample per class to support aggregation. This dataset is fed to the local models to extract embeddings. Then, the server calculates the norm of the Gram Matrix of the embeddings for each local model. The norm serves as an indicator of each model's inter-class separation capability in the embedding space. FedGraM identifies and removes potentially malicious models by filtering out those with the largest norms, then averages the remaining local models to form the global model. We conduct extensive experiments to evaluate the performance of FedGraM. Our empirical results show that with limited data samples used to construct the auxiliary dataset, FedGraM achieves exceptional performance, outperforming state-of-the-art defense methods.
\end{abstract}

\section{Introduction}

Federated Learning (FL) has gained significant traction for its ability to enable distributed machine learning without direct data sharing. This offers improved privacy, reduced communication costs, and scalability~\cite{fedavg,fedprox,scaffold,flapplication}. However, FL's decentralized nature introduces vulnerabilities~\cite{backdoorcritical,badnet,dba,adversariallen,targeted2,targeted3}, making it susceptible to untargeted attacks~\cite{lie,fang,shejwalkar2022back}. Attackers may infiltrate the system by posing as legitimate clients or hijacking existing clients, aiming to disrupt the training process and degrade the global model's generalization performance on the underlying data distribution.

Several studies have proposed robust aggregation methods to defend against untargeted attacks in FL~\cite{fltrust,fang,flmjr}. Most of these approaches aim to improve the FL system’s resilience to adversarial impact, ensuring the global model remains accurate even when some clients are malicious and submit manipulated models~\cite{trimean,crfl}. However, these methods often fall short in practical scenarios. Due to data heterogeneity, which introduces discrepancies among clients' data distribution, the benign models drift from each other, leading to worse convergence of the global model and more susceptibility to attacks from malicious clients. Consequently, the global model tends to achieve only sub-optimal performance under untargeted attacks.

In this paper, we aim to detect and remove malicious clients and fundamentally mitigate the impact of such attacks. We posit that the primary distinction between malicious and benign models lies in their contribution: benign models enhance the generalization of the global model, while malicious models degrade it. Thus, an intuitive approach to identify malicious models is to assess their accuracy on test data to evaluate generalization performance. However, maintaining a large dataset on the server is impractical due to privacy concerns. With only limited data, the estimated accuracy is often insufficient to capture the generalization differences between local models, particularly in the early stages of training. This constraint raises a critical challenge: \textbf{How can we estimate the generalization performance of local models with limited data?} Addressing this challenge is essential for advancing defense strategies in FL.

\begin{figure}[t]
    \centering
    \includegraphics[width=0.95\columnwidth]{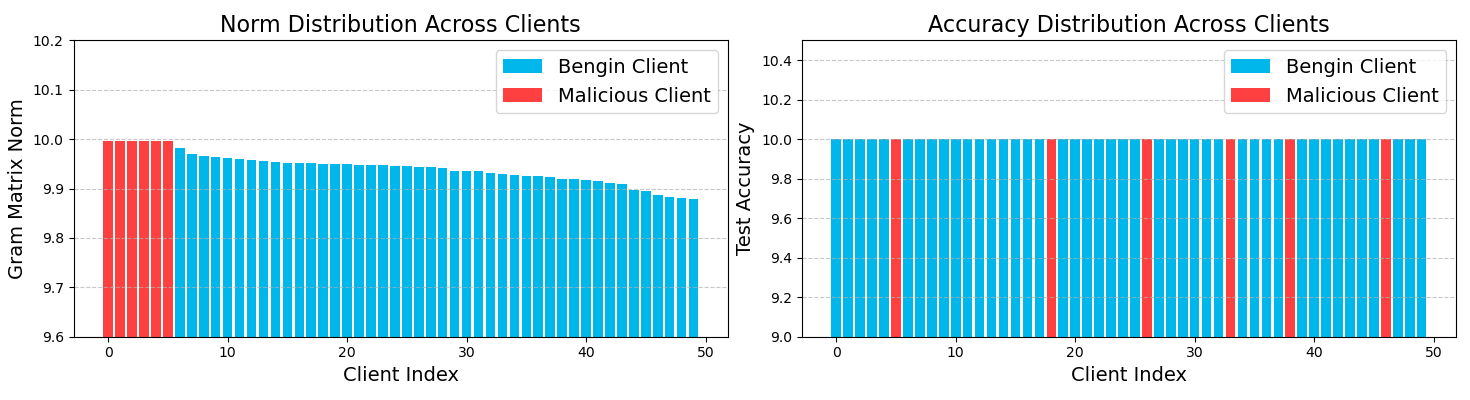}
    \caption{
    The norm distribution and the accuracy distribution across clients at the beginning stage of the training. The Gram matrix norm is calculated as we shown in Section \ref{sec:method} with only 10 data samples. The accuracy is estimated with 100 data samples. The detailed setting of this demo experiment are refereed to Section \ref{supp_demoexpsetting} in appendix. In each figure, we arrange the clients in descending order based on the corresponding values. Accordingly, with limited data, the Gram matrix norm is a better indicator to capture the generalization divergence between local models.}
    \label{introduction}
\end{figure}

To overcome this limitation, in this paper, we explore an alternative way of estimating generalization. In deep learning models, a key function of the representation layers is to separate data belonging to different classes within the embedding space, facilitating decision-making in subsequent layers. Compared with accuracy, the divergence between embeddings across classes can be a better estimator of generalization as it provide a more fine-grained indicator with less data required. We show a demo experiment result in Figure \ref{introduction} to demonstrate our insight.

In light of this, we propose a novel robust aggregation named FedGraM to detect and remove the malicious clients. In FedGraM, the server maintains an auxiliary dataset containing only one data sample per class. For each local model, the server inputs the dataset to extract the embeddings through the model's representation layers. After that, the server normalizes the embeddings and calculates the product of embeddings and its transpose matrix to obtain the Gram matrix. The norm of the Gram Matrix captures the inter-class separation ability of the representation layers, reflecting the generalization property of the model. Subsequently, the server discards the local models with the highest norms which are possible to be the malicious models, averaging the remaining local models to generate the global model. We conduct comprehensive evaluations on FedGraM. According to our experiment results, FedGraM is effective in defending against untargeted attacks with few data samples maintained by the server. It outperforms state-of-the-art defense methods. 

In summary, we make the following contributions:
\begin{itemize}
    \item We explore how the behavior of a deep learning model's representation layers in separating data representing different classes can serve as a reliable estimate of generalization in data-limited situations.

    \item We propose FedGraM, a robust aggregation method for defending against untargeted attacks in FL. It leverages the embedding Gram Matrix to indicate the generalization performance of local models. It filters out the local models with the worst generalization performance which have the potential to be malicious models

    \item We conduct extensive experiments to evaluate FedGraM. Our empirical results demonstrate that, with only limited data samples maintained by the server, FedGraM is effective in defending against untargeted attacks and outperforms state-of-the-art defense methods.

\end{itemize}

\section{Related Work}

\subsection{Untargeted Attack}

A body of research has focused on studying untargeted attacks in FL~\cite{cao2022mpaf,xie2024model}. These attacks, which aim to prevent model convergence and threaten the model's utility on the underlying data distribution, have garnered significant attention in the research community. According to the capability of the adversaries, the untargeted attacks can be categorized as model poisoning attacks and data poisoning attacks. Regarding data poisoning attacks\cite{fang,shejwalkar2022back},malicious clients set up the wrong label to the data to manipulate the label distribution. Regarding model poisoning attacks\cite{lie,fang,ndss,mpaf}, malicious clients directly manipulate the local models to prevent the convergence of the global model.

\subsection{Robust Aggregation}

A line of works proposes robust aggregation to defend against untargeted attacks~\cite{fedredefense,fldetector,baffle,bucket,foundationfl,rfa}. A major direction of defense is to perform dimension-wised strategy to alleviate the impact of malicious models\cite{trimean,rlr}. Another direction of works consider the model parameter as a lone vector and setup strategy to filter malicious models\cite{mmkrum,bulyan,flame}. There are also methods bounding the norm of model to guarantee the robustness\cite{crfl,naseri2020local}. Some method setup auxiliary dataset on the server to provide trusted information to support defenses\cite{fang,fltrust}. Our work follows this assumption and relaxes the requirement of the auxiliary dataset.

\section{Background}

\subsection{Federated Learning}

We consider FL is conducted for classification tasks in which the data will be classified into totally $K$ classes. The goal of FL is to solve the following problem:
\begin{equation}
    \underset{w}{\min }\left \{    J(w)=\frac{1}{N}\sum_{i=1}^{N}J_{i}(w)  \right \}  
\end{equation}
Where $w$ represents the model, and $J(w)$ denotes the global objective function, defined as the average of the local objective function $J_{i}(w)$ across all clients. $N$ is the number of the total clients. Further, we define the local objective of clients as follows:

\begin{equation}
J_{i}(w)=\underset{(x,y)\sim D_{i}  }{\mathbb{E}}\left [ L(F(w;x),y) \right ] 
\end{equation}

The local objective function for each client is defined as the expectation of loss function over its corresponding local dataset. $D_{i}$ is the local dataset owned by the $i$-th client. $x$ and $y$ represent the feature and the class of a data point randomly sampled from the local dataset. The loss function, denoted by $L$, is chosen as Cross Entropy in this paper to evaluate classification performance. $F(w;x)$ is defined as the function mapping the input feature $x$ to the output, parameterized by the model $w$.

In general, the deep learning models consist of two main components:(1) representation layers which transform input from the original feature space to an embedding space, and (2) decision layers, which generate a classification decision based on the embeddings for a given learning task. In this paper, we decompose the model $w$ into two parts. $\phi$ denotes the representation layers and $v$ denotes the decision layers. Further, we define $F(w;x)$ as follows:

\begin{equation}
F(w;x)=g(v;f(\phi;x))
\end{equation}

Where $f$ is the function that transforms the input from feature space to the embedding space via the representation layers $\phi$. $g$ is the decision function that maps the embeddings to the final results through the decision layers $v$.

Generally, FL iteratively performs the following steps to solve the problem:
\begin{itemize}
    \item \textbf{Step 1.} The server samples a subset of clients and broadcasts the current global model to these clients.
    \item \textbf{Step 2.} After receiving the global model, the sampled clients initialize their local models as the global model. The sampled clients perform a fixed step of stochastic gradient descent to update their local models towards the local objectives. When local training is finished, the clients send the local model back to the server.
    \item \textbf{Step 3.} The server collects the local model from the sampled clients. Then the server aggregates the local model based on the aggregation algorithm to generate the global model for the next round of training.
\end{itemize}

\subsection{Threat Model}

In this paper, we focus on the scenario in which the FL system faces security threats from malicious clients.  The goal of the malicious clients is to degrade the global model’s generalization performance on the underlying data distribution, ultimately reducing its accuracy. To achieve this, they compromise a subset of clients, either by injecting malicious data to perform data poisoning attacks or by manipulating the updates of the uploaded models to conduct model poisoning attacks. The remaining clients are benign and participate in local training honestly to support the FL process. The attackers are colluding and upload malicious models in each communication round to perform attack. The central server is responsible for defending against these adversarial attacks, typically employing robust aggregation to mitigate malicious clients' impact. In the rest of this paper, we refer to the local models uploaded by malicious clients as malicious models and those uploaded by benign clients as benign models.

\section{Methodology}
\label{sec:method}

\begin{figure*}[t]
    \centering
    \includegraphics[width=0.9\columnwidth]{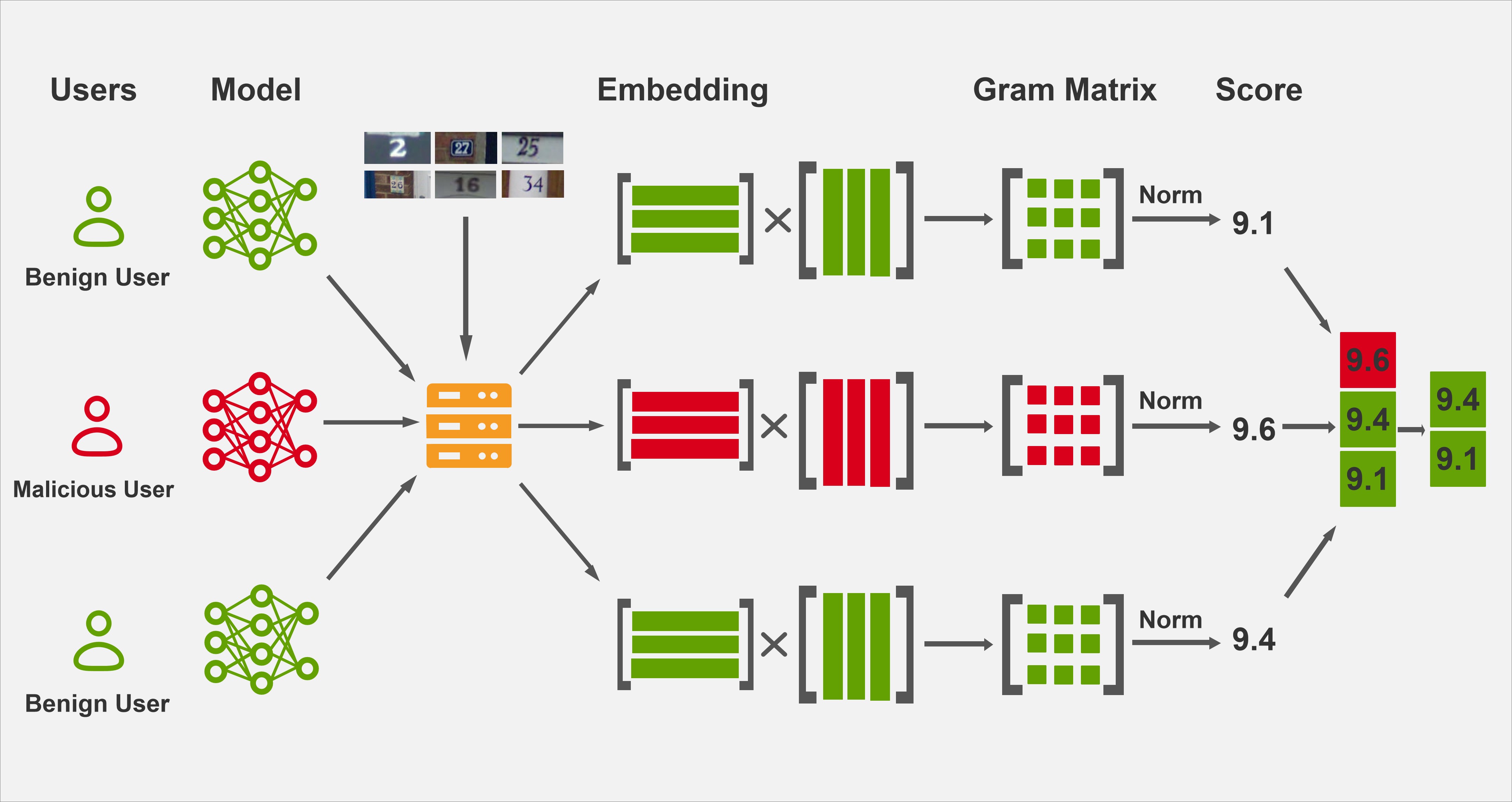}
    \caption{The overview of FedGraM. For each uploaded local model, the server feeds the auxiliary dataset to its representation layers to extract the corresponding embeddings. The server then calculates the product of the normalized embedding matrix and its transpose to get Gram Matrix. After that, the server calculates the norm of each Gram Matrix and removes the model with the highest norm. At the end, the server averages the remaining models to generate the global model.}
    \label{FedGraM}
\end{figure*}

\subsection{FedGraM}

FedGraM is a robust aggregation method which supposed to be utilized by the server to defend against untargeted attacks. In FedGraM, the server maintains an auxiliary dataset to support aggregation. We denote this dataset as $D_{s}$. Considering FL is conducted to solve a $K$-classes classification task, $D_{s}$ contains $K$ data samples, with exactly one data sample for each class. In each communication round, for each local model received by the server, the server feeds $D_{s}$ to the model's representation layers and obtains the embeddings. The server concatenates $K$ embeddings into an embedding matrix. Specifically, for the $i$-th client, its embedding matrix is presented as :

\begin{equation}
    p_{i,j}=f(\phi_{i};x_{j})
\end{equation}

\begin{equation}
    P_{i}=\left [ p_{i,0},p_{i,1}...p_{i,K-1} \right ] ^{T}
\end{equation}
Where $x_{j}$ represents the feature of the data sample with class $j$ in $D_{s}$. $\phi_{i}$ denotes the representation layers of the $i$-th client's local model. $p_{i,j}$ represents the embedding of $x_{j}$ on the $i$-th client's local model. $P_{i}$ represents the embedding matrix of the $i$-th client. 

Further, we normalize the embedding matrix so that the Euclidean norm of each row vector equals to $1$. After that, we calculate the Gram Matrix of the embeddings. The calculation is formed as:
\begin{equation}
    G_{i}=P_{i}P_{i}^{T}
\end{equation}
Where $G_{i}$ is the Gram Matrix for the $i$-th client. It is calculated as the product of embedding matrix $p_{i}$ and its transposed matrix.

Then, the server calculates the Euclidean norm of $G_{i}$ and records it as the score for the local model of the $i$-th client. After calculating all the scores for uploaded local models, the server eliminates the local models with the highest $C$ scores which have the potential to be malicious models. Then the server calculates the average of the rest local models as the global model in this communication round. We have shown the procedure of FedGraM in Figure \ref{FedGraM}. In our empirical evaluation, we set $C = 30\%$. In Section \ref{sec:ablation study}, we have conducted extensive experiments to support our setting and investigate the impact of $C$.

\subsection{Why FedGraM works}

In untargeted attacks, the goal of the malicious clients is to degrade the accuracy of the global model on underlying data distribution. As they are intended to destroy the generalization of the global model, the uploaded malicious models are supposed to have a bad generalization performance and do not contribute to the generalization of the global model. Considering a deep learning model, inter-class separation is the fundamental function of its representation layers, which supports the decision-making of the following layers. Within embedding space, these layers cluster data from the same class while separating data from the different classes, thereby enhancing the classification and guaranteeing generalization. \textbf{Therefore, the behavior of the representation layers can serve as an indicator of generalization performance and help identify the malicious models.}

Regarding FedGraM, we consider the Gram Matrix of a certain uploaded local model. For convenience, in this explanation, we drop the subscript of the index of the client. The element in the $x$-th row and $y$-th column of the Gram Matrix $G$ is formed as :
\begin{equation}
    G_{x,y}=\left \langle p_{x},p_{y} \right \rangle = \left \| p_{x} \right \|_{2} \cdot  \left \| p_{y} \right \|_{2} \cdot \cos (\theta )
\end{equation}
Where $G_{x,y}$ denotes the element in the $x$-th row and 
$y$-th column of the matrix $G$. $\theta$ is the angle between $p_{x}$ and $p_{y}$. Since we have done normalization on the embeddings before the calculation of Gram Matrix, $\left \| p_{y} \right \|_{2}$ and $\left \| p_{y} \right \|_{2}$ always equal to $1$. Therefore, $G_{x,y}$ can be simply formed as:
\begin{equation}
    G_{x,y}= \cos (\theta )
\end{equation}
Accordingly, $G_{x,y}$ represents the cosine similarity between the embedding of data with class $x$ and class $y$. When $x=y$, $G_{x,y}$ lies on the diagonal of the matrix and is always equal to 1. When $x\ne y$, the off-diagonal elements of $G$ indicate the local model's ability to separate embeddings with class $x$ and class $y$. A lower value of $G_{x,y}$ reflects better separation between these classes. \textbf{Thus, the Gram Matrix captures the ability of the representation layers to achieve inter-class separation within the embedding space.}

In conclusion, FedGraM measures the norm of the Gram Matrix to estimate the generalization performance of each uploaded model. A higher norm indicates that the model has a worse generalization performance. Hence, FedGraM removes the models with the highest norm which are potential to be the malicious models, and averages the remaining models to aggregate the global model.

\subsection{Auxiliary Dataset}

In the general framework of FL, the server typically does not have access to any data, which poses a challenge for FedGraM's requirement of an auxiliary dataset. To address this, we consider two solutions. First, when data owners collaborate in FL, one of them could act as the server and collect a small auxiliary dataset to enable FedGraM. Second, the server could encourage all clients to contribute to a shared, public auxiliary dataset for FedGraM. This dataset has no specific quality requirements; ideally, it includes one randomly selected data sample per class. However, in both approaches, the auxiliary dataset may not cover all classes. We evaluate FedGraM's performance under this limitation in Section \ref{sec:ablation study}.

\section{Empirical Evaluation}

\subsection{Experiment Setting}

We introduce a default setting for our experiments; some settings may change in certain experiments. The FL system involves $500$ clients with $10\%$ of clients are malicious clients. In each communication round, $10\%$ of the clients are randomly selected to participate in the training. We used the Dirichlet distribution $Dir(\beta)$ to simulate the data heterogeneity in FL. The experiments were conducted on the CIFAR10\cite{cifar10} dataset, SVHN\cite{svhn} dataset, and CIFAR100\cite{cifar10} dataset. The auxiliary dataset $D_s$ is randomly sampled from the union of the clients’ local dataset and is excluded from the clients' local training. We record the highest test accuracy achieved by the global model to reflect the performance of the methods. The detials of the experiment settings are referred to Section \ref{supp_experiment_setting} in appendix.

\subsection{Comparison with SOTA defenses}

We compare the defending performance of FedGraM with SOTA defense methods. We concentrate on the performance with different data heterogeneity and with different ratio of malicious clients. We implement attacks including LIE\cite{lie}, Fang\cite{fang}, MinMax\cite{ndss}, MinSum\cite{ndss}, MPAF\cite{mpaf}, Label Flip\cite{fang} and Dynamic Label Flip\cite{shejwalkar2022back} to evaluate the performance of defense methods.
We implement defense methods including Trimean\cite{trimean}, Norm Bound\cite{naseri2020local}, CRFL\cite{crfl}, FLTrust\cite{fltrust}, FLAME\cite{flame}, RONI\cite{fang}, Bucket\cite{bucket}, FedRola\cite{fedrola}, Krum\cite{mmkrum}, Bulyan\cite{bulyan}, FoundationFL\cite{foundationfl}, RFA\cite{rfa}, RLR\cite{rlr} for comparison. The detailed introduction of attacks and defenses are refereed to Section \ref{supp_evaluated_attacks} and Section \ref{supp_evaluated_defenses} in appendix.

\textbf{Different Data Heterogeneity} In this experiment, we evaluate the performance of defense methods with different data heterogeneity. Specifically, we set $\beta \in \{10,1,0.2\} $ to simulate different degrees of data heterogeneity among clients. Due to page limitations, we only show the performance of partial defenses in Figure \ref{Comparison}. The entire experiment results are referred to Section \ref{supp_comparison_noniid} in appendix. Accordingly, FedGraM achieves high accuracy under all attacks in all situations which demonstrates the effectiveness of FedGraM in defending untargeted attacks. The vulnerability of the attacks is revealed in their malicious behavior in the embedding space which can be captured by FedGraM. Therefore, FedGraM can correctly detect and remove malicious models from local models. Meanwhile, the existing defense methods do not achieve satisfactory performance for certain attacks. Due to data heterogeneity and cross-device scenarios, they may not retain robustness in the experiments. We also conduct experiments in cross-silo scenario in which we setup the FL system with different clients number, different participate ratio and different model architecture. The experiment results in cross-silo is shown in Section \ref{supp_comparison_silo} in appendix which is also demonstrating the effectiveness and superiority of FedGraM.

\begin{figure*}[h]
    \centering
    \includegraphics[width=0.95\columnwidth]{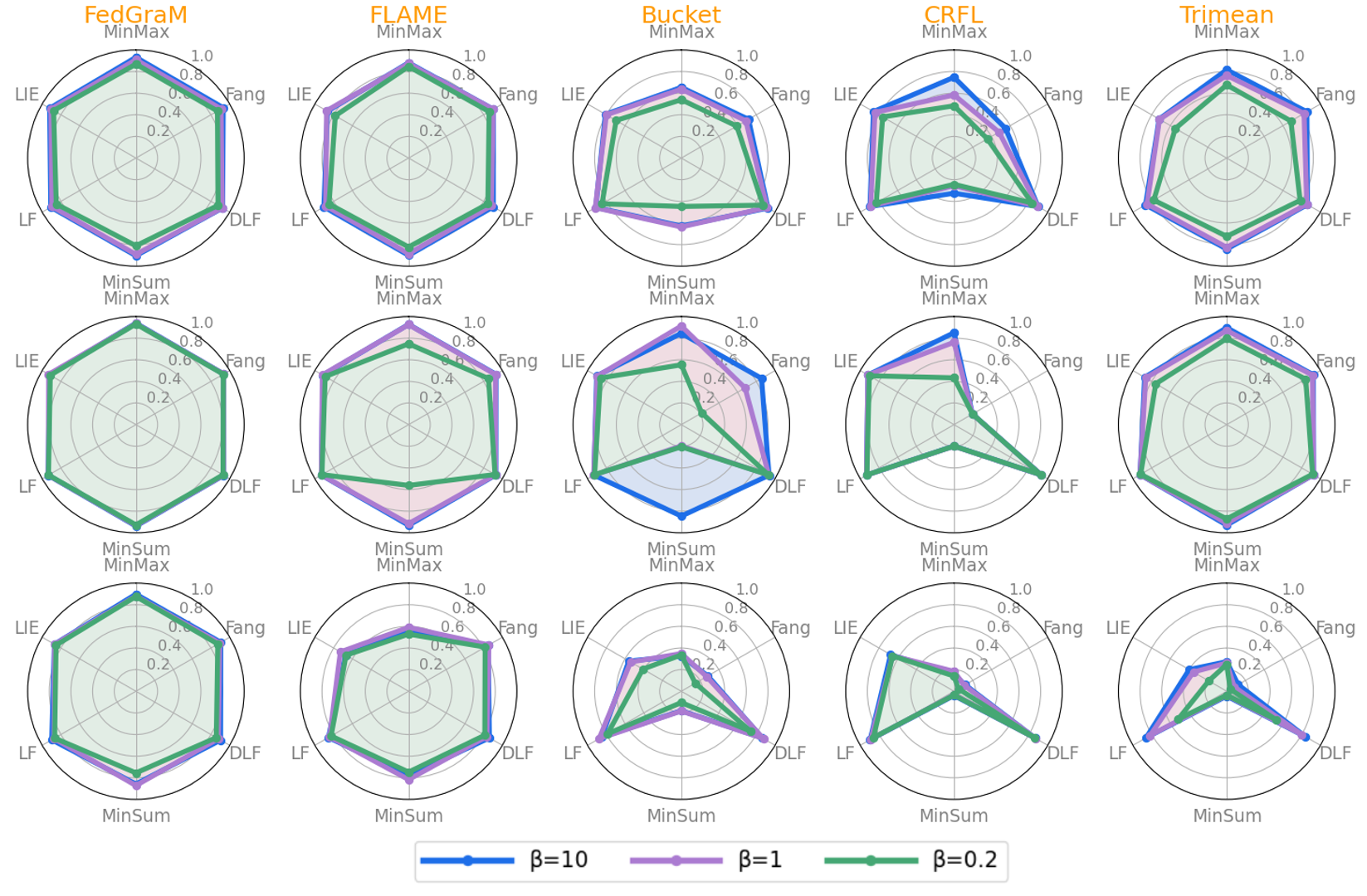}
    \caption{The experiment results of the comparison between FedGraM and SOTA defenses. We draw a spider chart for each defense to represent its performance in defending against untargeted attacks. Each dimension of the map represents the performance of the corresponding defense in defending a certain type of attack. From the top row to the bottom row, we present the results in CIFAR10, SVHN, and CIFAR100.}
    \label{Comparison}
\end{figure*}
\begin{table}[h]
\centering
\caption{The experiment results of the impact of malicious clients ratio in CIFAR10 dataset. We show the performance(Test accuracy (\%)) of FedGraM and existing defense methods with different malicious clients ratios.   }
\label{Attackratio}
\begin{tabular}{cccccccccc}
\toprule
Attacks & \multicolumn{3}{c}{LIE} & \multicolumn{3}{c}{Fang} & \multicolumn{3}{c}{MinMax} \\ \hline
Ratio   & 15\%   & 10\%   & 5\%   & 15\%   & 10\%   & 5\%    & 15\%    & 10\%    & 5\%    \\ \bottomrule
Trimean & 48.76  & 57.31  & 65.61 & 57.54  & 65.98  & 70.40  & 50.77   & 61.12   & 70.74  \\
CRFL    & 62.65  & 67.49  & 71.56 & 28.19  & 38.37  & 67.59  & 32.67   & 46.70   & 68.09  \\
RONI    & 62.37  & 67.87  & 72.84 & 70.36  & 74.40  & 74.21  & 48.52   & 64.67   & 70.61  \\
FedGraM & 72.57  & 72.47  & 73.91 & 73.94  & 73.72  & 74.16  & 73.06   & 72.72   & 72.98  \\ \bottomrule
\end{tabular}

\end{table}

\textbf{Different Malicious Clients Ratio} We further explore the impact of the ratio of malicious clients. We set the ratio of malicious clients to $5\%$, $10\%$, and $15\%$. The $\beta$ are set to $1$. The results are shown in Table \ref{Attackratio}. The entire version of the results are shown in Section \ref{supp_comparison_attackraio} in appendix.  Consequently, the performance of FedGraM is consistent with the ratio of malicious clients. In most situations, it performs best among the evaluated defense methods. More importantly, it is obvious that the performance of other defense methods degrades as the ratio of malicious clients increases. On the contrary, the performance of FedGraM is not impacted by the ratio of malicious clients. However, in some situations, FedGraM is worse than other defense methods. In most of these situations, the FedGraM achieves a similar performance as the best method which can also demonstrate its effectiveness. 

\subsection{Ablation Study}
\label{sec:ablation study}
\begin{figure*}[h]

    \centering
    \includegraphics[width=1.0\columnwidth]{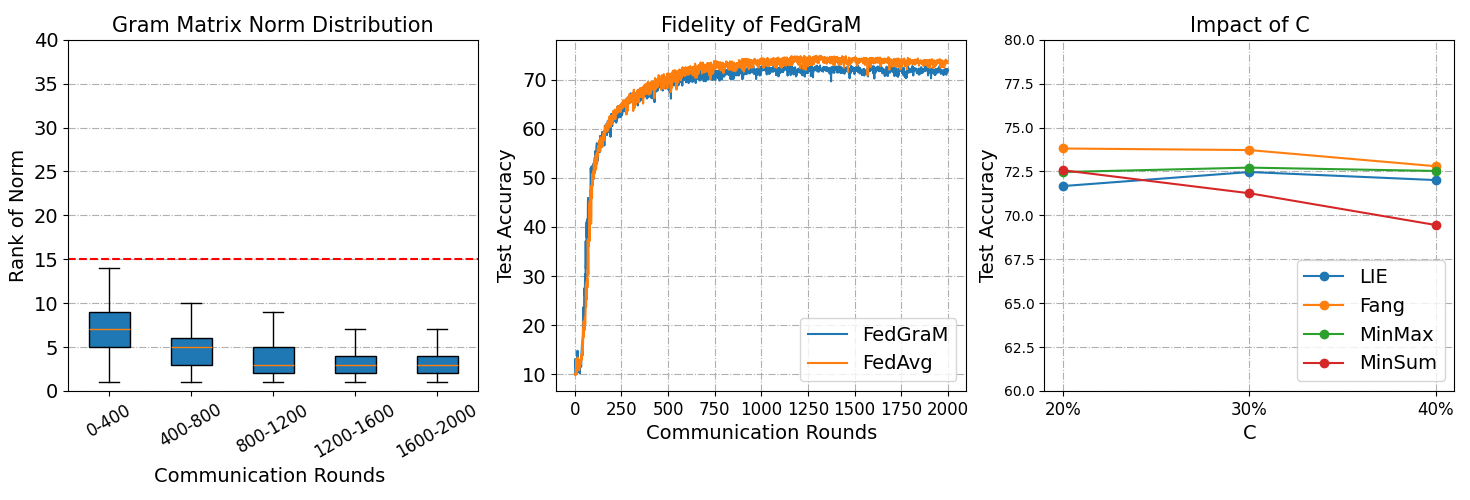}
   
    \caption{From lest to right, we show the experiment results of Gram matrix norm distribution, Fidelity of FedGraM and Impact of C.}
     \label{ablationstudy_fig}
\end{figure*}

\begin{figure*}[h]
    \centering
    \includegraphics[width=1.0\columnwidth]{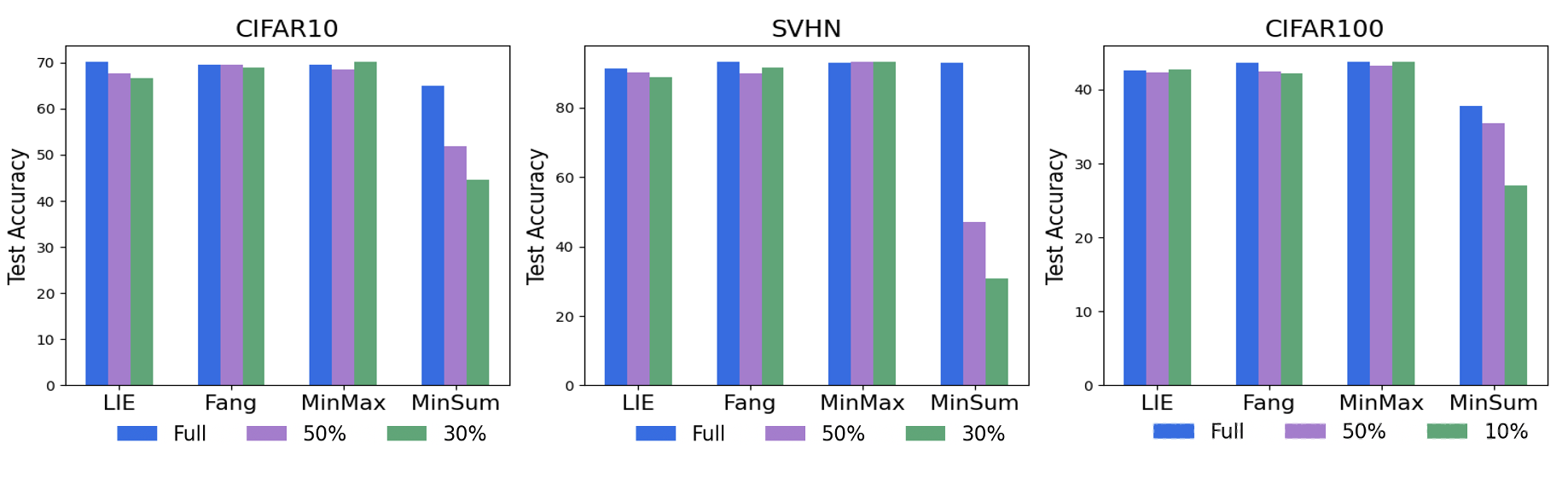}
    \caption{The experiment results of the impact of the quality of the auxiliary dataset on the performance of FedGraM.}
    \label{Quality}
\end{figure*}

The hyperparameter $C$ and the auxiliary dataset $D_s$ are two crucial factors which determine the effectiveness of FedGraM. We conduct comprehensive evaluations to investigate their impact and support our hyperparameter setting.

\textbf{Gram Matrix Norm Distribution} To begin with, we show the distribution of the norm of GraM matrix across clients during the training process. The experiment in conducted under LIE attack with $\beta = 0.2$ in CIFAR10. We show the rank of malicious clients' norms among all the local models in Figure \ref{ablationstudy_fig}. The norm are ranked in descending order from bottom to top. The norm of malicious models always distributed at the largest value which demonstrate the effectiveness of the FedGraM in distinguishing malicious clients. During the training, the lowest norm of malicious models is ranked below 30\% of the total norm. Therefore, we set $C=30\%$ and filter the local models with the highest $30\%$ norm in each round.

\textbf{Fidelity of FedGraM} With $C=30\%$, we evaluate the fidelity of FedGraM. Specifically, we estimate the test accuracy without any attack to reveal the performance of FedGraM and compare it with FedAvg. The experiment are conducted in CIFAR10 with $\beta = 1$ and the result is shown in Figure \ref{ablationstudy_fig}. The experiment results in more dataset with other situation of data heterogeneity is shown in Section \ref{supp_fidelity} in appendix. Accordingly, the similar accuracy curves have demonstrated that FedGraM does not sacrifice much utility of the model for robustness. A potential reason for this is that filter a part of local models can be treated as setting a lower client sample rate in each communication round which would bring significant convergence and generalization loss demonstrated by \cite{flconvergence} . 

\textbf{Impact of $C$} Further, we investigate the impact of $C$ to the defending performance of FedGraM. We set $C \in \{20\%,30\%,40\%\}$. We set $\beta = 1$. The results in CIFAR10 dataset are shown in Figure \ref{ablationstudy_fig}. We show an entire version of results in Section \ref{supp_impactofc} in appendix.  Accordingly, while $C=40\%$, an excessive number of local models were removed which led to performance degradation in all datasets. FedGraM has shown similar performances with $C=20\%$ and $C=30\%$. However, under certain situations, $C=20\%$ is insufficient to defend against attacks. As a result, setting $C=30\%$ is the safest choice to guarantee robustness. Although there might not be that many malicious clients in the FL system, appropriately removing some local models can mitigate potential threats. Consequently, our experiment results support our setting of $C = 30\%$.

\textbf{Quality of Auxiliary dataset} Finally, we investigate the impact of the quality of the auxiliary dataset maintained by the server on the performance of FedGraM. We consider the practical scenario in which the server only has a subset of the classes. Specifically, for CIFAR10 and SVHN, we simulate the situations in which the auxiliary dataset includes full labels, $50\%$ labels, and $30\%$ labels. For CIFAR100,  we simulate the situations in which the auxiliary dataset includes full labels, $50\%$ labels, and $10\%$ labels. We set $\beta=0.2$. According to the results shown in Figure \ref{Quality}, FedGraM can retain robustness with a subset of classes under certain attacks. However, it is obvious that without the integrity of class distribution, FedGraM is insufficient to defend against MinSum attack in all datasets. A potential reason is that, with incomplete classes, FedGraM cannot capture the entire behavior of the local models in the embedding space, leading to its poor performance in distinguishing malicious models. In practical application scenarios, we encourage the server to collect the full classes of data as much as possible to ensure robustness.

\section{Limitation}

Given sufficient knowledge of FedGraM, malicious clients may attempt to deploy adaptive attacks to bypass its defense mechanisms. In this section, we analyze this potential threat and evaluate the robustness of FedGraM against such adaptive attacks.

Ideally, the adaptive attack covers two property. First, the malicious model should output random embeddings for all input data to evade detection by FedGraM. Secondly, the embeddings of data within the same class should also be orthogonal, thereby destroying any meaningful clustering in the embedding space. Inspired by~\cite{wang2020understanding}, we design an adaptive attack in which malicious clients train their local models to enforce pure uniformity in the embedding space, entirely disregarding the objective of correct classification. Specifically, malicious clients follow the same local training procedure as benign clients but replace the standard Cross Entropy loss with the following loss function:

\begin{equation}
    L(w)=\log_{}{}  \underset{x_1,x_2\sim D_{i}  }{\mathbb{E}}[e^{-\left \| f(\phi;x_1)-f(\phi;x_2) \right \|_{2}^{2} }]
\end{equation}

where the model is optimized to maximize the distance between embeddings of any pair of data points, thereby achieving pure uniformity in the embedding space.

\begin{wraptable}{r}{8cm}
\centering
\renewcommand{\arraystretch}{1.2}
\caption{Performance of FedGraM under adaptive attacks.}
\begin{tabular}{cccc}
\toprule
Dataset                   & $\beta$ & FedGraM-AVG & FedGraM-Trim \\ \bottomrule
\multirow{2}{*}{CIFAR10}  & 10   & 22.04\%     & 72.25\%      \\
                          & 0.2  & 40.50\%     & 65.77\%      \\
\multirow{2}{*}{SVHN}     & 10   & 34.95\%     & 93.20\%      \\
                          & 0.2  & 43.63\%     & 92.14\%      \\
\multirow{2}{*}{CIFAR100} & 10   & 5.39\%      & 42.65\%      \\
                          & 0.2  & 11.71\%     & 37.82\%      \\ \bottomrule
\end{tabular}

\label{Adaptive}
\end{wraptable}

Furthermore, we evaluate the performance of FedGraM under adaptive attacks. Since FedGraM functions as a detection mechanism, it can be combined with any aggregation method to enhance the overall robustness of the system. To this end, we employ FedAvg and Trimean as aggregation methods following FedGraM's detection phase. The experimental results are summarized in Table \ref{Adaptive}. Our findings indicate that adaptive attacks indeed undermine the robustness of FedGraM to some extent. However, while adaptive attacks enable malicious models to evade detection by FedGraM, they simultaneously amplify the divergence between malicious and benign models. This divergence makes the impact of adaptive attacks more detectable and mitigable through classical statistical-based robust aggregation methods. Specifically, when Trimean is applied after FedGraM detection, the global model's accuracy remains largely unaffected, demonstrating the resilience of this combined approach. In conclusion, although adaptive attacks pose a challenge to FedGraM's detection capabilities, their impact can be effectively mitigated by integrating robust aggregation methods. This suggests that even if malicious models bypass detection, the system can still maintain its robustness through subsequent statistical-based aggregation techniques.

\section{Conclusion}

We propose a novel robust aggregation method named FedGraM to defend against untargeted attacks in FL. In FedGraM, the server maintains an auxiliary dataset to support the aggregation. In each communication round, the server feeds the dataset to received local models to extract the corresponding embeddings. The server multiplies the embedding matrix and its transpose matrix to obtain the Gram Matrix. The norm of Gram Matrix captures the capability of the local model's representation layers in inter-class separation in embedding space. It is an important property supporting the generalization of the deep learning model. The server filters out the local models with the highest norm of Gram Matrix which are potential to be the malicious models. The remaining local models are averaged to generate the global model. We have conducted extensive experiments to evaluate the performance of FedGraM. As our empirical results show, FedGraM is effective in defending against untargeted attacks with limited data samples on the server. It is comprehensive to defend all the evaluated attacks and outperforms SOTA defense methods. 

\bibliographystyle{plain}
\bibliography{main.bib}

\clearpage
\appendix

\section{Additional Experiment Setting}

\subsection{Evaluated Attacks}
\label{supp_evaluated_attacks}

\subsubsection{LIE}

LIE\cite{lie} is a common untargeted attack that aims to prevent the convergence of the global model. Adversaries set up a desired aggregated direction on parameter space which is inverse to the direction of convergence. By crafting values between the mean and the desired direction, the malicious model appears closer to the mean than some benign models that are at the extremes of the distribution. This allows attackers to bypass defense methods and prevent the convergence of the global model. Formally, the attacks can be expressed as: 
\begin{equation}
    z^{max}=max_{z}(\phi(z)< \frac{n-m-s}{n-m}  ),s=\left \lfloor \frac{n}{2}+1  \right \rfloor-m 
\end{equation}
\begin{equation}
    w^{m}_{i,j}=mean(w^{b}_{i,j})-z^{max}\cdot std(w^{b}_{i,j})
\end{equation}
Where $n$ and $m$ represent the number of total clients and malicious clients respectively. $\phi(z)$ is the Cumulative Standard Normal Function. $w^{b}_{i,j}$ and $w^{m}_{i,j}$ are the $j$-th parameter of the local updates of the $i$-th benign client and malicious client respectively. The mean and the standard deviation of benign clients can be captured by malicious clients leveraging some hijack tools. They can also be simulated by the malicious clients themselves.

\subsubsection{Fang}
Fang\cite{fang} is a model poisoning attack where adversaries manipulate local updates to steer the global model towards the inverse direction of convergence. Fang is designed with specific attacks tailored to different aggregation algorithms to ensure their stealthiness. Empirical results demonstrate that attacks designed for certain aggregation algorithms can be transferred to others with minimal loss of utility. Therefore, in this paper, we employ the Fang attack tailored for the Trimmed Mean aggregation.

Specifically, considering the $j$-th dimension of the model parameters, $s_j$ is set up to represent the changing direction of the global model. $s_j=1$(or$s_j=-1$) means that the $j$-th dimension of the parameter increases(decreases) upon the previous iteration. $w_{max,j}$ and $w_{min,j}$ denote the maximum and minimum of the $j$-th dimension of the local updates on benign clients, i.e., $w_{max,j}=max \left \{   w^{b}_{1,j},w^{b}_{2,j},...,w^{b}_{m,j}\right \}$ and $w_{min,j}=min \left \{   w^{b}_{1,j},w^{b}_{2,j},...,w^{b}_{m,j}\right \}$ The $j$-th dimension of the malicious model is formed as:
\begin{equation}
    w^{m}_{i,j}\in \left [ w_{max,j},b\cdot w_{max,j} \right ] ( s_j=-1,w_{max,j}>0)
\end{equation}
\begin{equation}
    w^{m}_{i,j}\in \left [ w_{max,j}, w_{max,j}/b \right ] ( s_j=-1,w_{max,j}\le0)
\end{equation}
\begin{equation}
    w^{m}_{i,j}\in \left [ w_{min,j}/b, w_{min,j} \right ] ( s_j=1,w_{min,j}>0)
\end{equation}
\begin{equation}
    w^{m}_{i,j}\in \left [ b\cdot w_{min,j}, w_{min,j} \right ] ( s_j=1,w_{min,j}\le0)
\end{equation}
Where the $j$-th dimension of the malicious model $w^{m}_{i,j}$ is randomly sampled in a fixed range determined by $w_{max,j}$, $w_{min,j}$ and $s_j$.
\subsubsection{MinMax \& MinSum}
MinMax and MinSum are model poisoning attacks proposed by \cite{ndss}. They formulate the objective of untargeted attacks as an optimization problem and crafts malicious models by solving this problem. MinMax and MinSum are designed to solve the following optimization problem to construct the malicious models:
\begin{equation}
    \underset{\gamma }{argmax} \,\,  \underset{i}{\max}\left \| w^{m}-w^{b}_i \right \|_{2}\le \underset{i,j}{\max}\left \| w^{b}_{i}-w^{b}_{j} \right \|_{2}    
\end{equation}

\begin{equation}
    \underset{\gamma }{argmax} \,\,  \sum_{i}\left \| w^{m}-w^{b}_i \right \|_{2}\le \underset{i,j}{\max}\left \| w^{b}_{i}-w^{b}_{j} \right \|_{2}    
\end{equation}

\begin{equation}
    w^{m}=w^{b}-\gamma \cdot w^{p} ,w^{p}=-\frac{w^{b}}{\left \| w^{b} \right \|_{2} },w^{b}=\underset{i}{f_{avg}(w^{b}_{i})} 
\end{equation}
$(16)$ is the objective of MinMax attack, and $(17)$ is the objective of MinSum attack. $f_{avg}$ denotes the standard aggregation algorithm FedAvg. These attack aim to generate a malicious update that is close to benign updates, and meanwhile, points in the inverse direction to the benign updates.

\subsubsection{Label Flip \& Dynamic Label Flip}

Label Flip\cite{fang} and Dynamic Label Flip\cite{shejwalkar2022back} are data poisoning attacks that manipulate the data set of the malicious clients to impact the performance of the global model. Specifically, Label Flip flips the label of each training instance. It flip a label $l$ as $L-l-1$, where L is the number of classes in the classification problem and $l=0,1,... ,L-1$. In Dynamic Label Flip, the malicious clients compute a surrogate model, using the available benign data, and flip the label to the least probable label.

\subsubsection{MPAF}
MPAF\cite{mpaf} introduces a novel FL attack method that does not require local data or relies on any system-level information. Unlike traditional poisoning attacks that depend on controlling real clients, MPAF effectively misguides the global model towards a poorly performing "baseline model" by injecting fake clients and constructing model updates in a specific direction. The effectiveness of this method lies in the fact that all fake clients consistently send updates towards the same target (the baseline model), ensuring high consistency. This results in the cumulative effect of the shift over multiple training rounds, making it difficult to eliminate, even in the presence of robust aggregation and pruning defense mechanisms. Consequently, this attack achieves a stable and potent poisoning effect.

\subsubsection{Scaling}
Scaling\cite{DBLP:journals/corr/abs-1807-00459} is a classic backdoor attack targeted at federated learning. Malicious clients reduplicate their local training data. A part of local data is embedded with triggers and assigned new labels. Subsequently, these clients train using both the vanilla data and the poisoned data. The resulting local model is further scaled to amplify its impact.

\subsubsection{DBA}

DBA\cite{dba} is a sophisticated adversarial strategy targeting federated learning systems, where multiple participants collaboratively train a shared machine learning model without sharing their raw data. Unlike traditional backdoor attacks that operate in centralized settings, DBA leverages the distributed nature of federated learning to implant backdoors into the global model. In this attack, malicious clients introduce poisoned data or model updates during local training, embedding hidden triggers that cause the global model to exhibit attacker-desired behaviors when specific patterns (e.g., certain pixel arrangements or keywords) are present. By distributing the backdoor task across multiple clients, DBA makes the attack more stealthy and harder to detect compared to centralized backdoor attacks. Defending against DBA requires robust aggregation techniques, anomaly detection mechanisms, and advanced privacy-preserving methods to ensure the integrity and security of federated learning systems.

\subsection{Evaluated Defenses}
\label{supp_evaluated_defenses}

\subsubsection{Trimean}

Trimmed Mean\cite{trimean} is an adaptation of the traditional Byzantine algorithm in FL. In Trimmed Mean, for each dimension of the global model, after receiving local updates from clients, the server excludes the largest and smallest $k$ values of the dimension and calculates the average of the remaining values to determine the dimension's result. While Trimmed Mean sacrifices some model utility, it has been empirically shown to effectively defend against basic poisoning attacks.

\subsubsection{Norm Bound}

Namely, Norm Bound aims to limit the behavior of the clients by bounding the vector norm of the local updates.
\cite{naseri2020local,shejwalkar2022back,sun2019can}
Specifically, Norm Bound can be treated as a variation of FedAvg since it induces a norm clipping before calculating the average of clients' local updates. It forms as:
\begin{equation}
    w=\frac{1}{n}\sum_{i=1}^{n}Clip(w_{i})  
\end{equation}
\begin{equation}
    Clip(w_{i})=w_{i}\cdot min(1,\frac{\left \| w_{i} \right \|_{2} }{p} )
\end{equation}
Accordingly, the norm of local updates is bounded by $p$. In practice, $p$ can be statically determined before learning or dynamically estimated during the training. The norm is usually estimated as Euclidean norm. Norm Bound is widely used in practice. It is efficient to defend against model poisoning attacks. The implementation of Norm Bound is simple as it can be combined with other cryptographic tools to enhance the privacy preservation of FL. 

\subsubsection{CRFL}
CRFL\cite{crfl} is an FL framework that focuses on ensuring robustness during both the training and inference phases. In CRFL, which operates on the server side, the framework computes the average of local updates similar to FedAvg. Moreover, it then clips the global model parameters to ensure their norm is bounded. Additionally, CRFL adds isotropic Gaussian noise directly to the aggregated global model parameters. Formally, the norm clipping and noise injection can be expressed as:
\begin{equation}
    Clip(w)=w\cdot min(1,\frac{\left \| w \right \|_{2} }{\rho} )
\end{equation}
\begin{equation}
    Perturb(w)=w+\epsilon ,\epsilon \sim \mathcal{N}(0,\sigma ^{2}\mathbf{\mathbb{I}}) 
\end{equation}
Where $w$ is the global model. $Clip$ bound the norm of global model $w$ by $\rho$. The noise added by $Perturb$ is sampled on Gaussian distribution $\mathcal{N}(0,\sigma ^{2}\mathbf{\mathbb{I}}) $. $\rho$ and $\sigma$ can be dynamically tuned during the training. During the inference phase, the server will smooth the final model with randomized parameter smoothing and make the final prediction based on the parameter-smoothed model. Theoretically, CRFL guarantees that the trained global model would be certifiably robust against the backdoor as long as the backdoor is within certain certified bounds.

\subsubsection{FLTrust}
In FLTrust\cite{fltrust}, the server itself collects a clean small training dataset (called root dataset) for the learning task and maintains a model (called server model) based on it to bootstrap trust. In each iteration, the server first assigns a trust score to each local model update from the clients, where a local model update has a lower trust score if its direction deviates more from the direction of the server model update. Then, the server the magnitudes of the local model updates such that they lie in the same hyper-sphere as the server model update in the vector space. 

\subsubsection{RONI}

In RONI\cite{fang}, we compute the impact of each local model on the error rate for the validation dataset and remove the local models that have large negative impact on the error rate. Specifically, suppose we have an aggregation rule. For each local model, we use the aggregation rule to compute a global model when the local model is included and a global model when the local model is excluded. We compute the error rates of the global models on the validation dataset. We define the error rate impact of a local model as the deviation between the accuracy of two global models. A larger error rate impact indicates that the local model increases the error rate more significantly if we include the local model when updating the global model. We remove the local models that have the largest error rate impact, and we aggregate the remaining local models to obtain an updated global model.

\subsubsection{FLAME}
FLAME\cite{flame} is a robust aggregation method. FLAME leverages HDBSCAN based on the cosine similarity between clients' local updates to cluster the local model and detect malicious models. It also conducts norm clipping and adds noise to the aggregation results to further enhance the robustness of the FL system.

\subsubsection{Bucket}

Bucket\cite{bucket} is a robust aggregation method. To aggregate the uploaded local models, Bucket splits the local models into several buckets. Each bucket computes the average of the local models in the bucket to represent the bucket. After that, the higher-level aggregation method is applied to the bucket models. In our experiment, we set the size of the bucket to $2$ and used Trimean as the aggregation method for bucket models.

\subsubsection{Krum}
Krum\cite{mmkrum} is an aggregation rule proposed to define against Byzantine attacks in FL. The objective of Krum is to select the most reliable gradient update from the gradient vectors submitted by $n$ clinets, thereby mitigating the impact of up to $f$ Byzantine nodes. To achieve this, Krum computes a score for each client's gradient update $w$,  which is defined as the sum of squared distances between its gradient update and those of its $n-f-2$ nearest neighboring clients. Subsequently, the client with the lowest score is chosen as the aggregation output. This score for the $i$-th client can be expressed as follows:
\begin{equation}
s(i) = \sum_{j \in \mathcal{N}_i} \| w_i - w_j \|^2,
\end{equation}
where $\mathcal{N}_i$ is the set of indices of the $n - f - 2$ nearest neighbors of $w_i$. In addition, this paper also proposed Multi-Krum, which is an extension of Krum, by selecting multiple reliable gradient updates instead of a single one to enhance the stability of the aggregation result.

\subsubsection{Bulyan}
Bulyan\cite{bulyan} is a Byzantine-resilient aggregation rule designed to address the security vulnerabilities inherent in high-dimensional spaces, where malicious clients may exploit the dimensionality curse to cause Stochastic Gradient Descent (SGD) to converge to ineffective suboptimal solutions. Bulyan builds upon an existing Byzantine-resilient aggregation rule (e.g., Krum, Brute, or GeoMed) and introduces a two-step process. First, the method iteratively applies the base aggreation rule to select $\theta = n - 2f$ gradient updates. In each iteration, the gradient updates closest to the current output of the base rule is identified, added to a selection set $\mathcal{S}_{s}$, and removed from the received set $\mathcal{S}_{r}$. This process continues until the size of $\mathcal{S}_{s}$ reaches $\theta$. Second, for each coordinate component $i$ of the gradient update, Bulyan computes the median of the $\theta$ selected gradients and averages the $\beta = \theta - 2f$ closest coordinates to this median, producing the final aggregated gradient. This coordinate-wise approach ensures that each component of the output is dominated by a majority of non-Byzantine contributions. The resulting gradient $G$ for the $i$-th coordinate can be expressed as follows:
\begin{equation}
G[i] = \frac{1}{\beta} \sum_{X \in M[i]} X[i],
\label{eq:gradient_aggregation}
\end{equation}
where $M[i]$ denotes the $\beta$-nearest gradients to the median in the $i$-th coordinate dimension. Bulyan significantly reduces the poisoning effect of Byzantine attacks through recursive selection and coordinate-level averaging.

\subsubsection{FedRola}
FedRoLA\cite{fedrola} addresses the vulnerability of FL to model poisoning attacks by proposing a layer-based aggreation defense mechanism. The core innovation of FedLoRA lies in leveraging the characteristics of DNN layers to detect malicious clients through similarity analysis while minimizing the false rejection rate of benign updates. First, FedLoRA dynamically identifies the most sensitive layers in the DNN for detecting malicious behavior. Then, for these selected layers, it introduces Layer Alignment Similarity Index (LASI) and Peer Consensus Similarity Index (PCSI) to analyze the anomaly of malicious updates at the layer-wise level. The LASI is derived by computing the cosine similarity between $i$-th client's $l$-th layer-wise updates $w_{i,l}^{t}$ and the global model's $l$-th updates:
\begin{equation}
    LASI_{i,l} = \frac{\langle w_{i,l}^{t}, \hat{w}_l^{t} \rangle}{\| w_{i,l}^{t} \| \cdot \| \hat{w}_l^{t} \|},
\end{equation}
where $t$ is the global communction round, $\langle \cdot \rangle$ and $\| \cdot \|$ respectively denotes the inner product and the edclidean norm. The PCSI is derived by computing the cosine similarity between $i$-th client's $l$-th layer-wise updates $w_{i,l}^{t}$ and other clinets' $l$-th updates:

\begin{equation}
    PCSI_{i,l} = \frac{1}{\lvert N^{t} \rvert - 1} \sum_{\substack{j \in N^{t} \\ j \neq i}} \frac{\langle w_{i,l}^{t}, w_{j,l}^{t} \rangle}{\| w_{i,l}^{t} \| \cdot \| w_{j,l}^{t} \|},
\end{equation}
where $N^{t}$ is the set of clinet at $t$ global communction round and $\lvert N^{t} \rvert$ is the number of clients. Additionally, FedRoLA employs a layer-wise weighted voting mechanism to compute a suspicion score for each client. If the score exceeds a predefined threshold, the client is flagged as potentially malicious and assigned a lower discount factor. Finally, the global aggregation is performed by adjusting the weights of client updates based on their respective discount factors:
\begin{equation}
    {w}^{t} = \sum_{i \in N^{t}} \alpha_i \cdot {w}_i^{t},
\end{equation}
where $\alpha_i$ the weight of $i$-th client. Through layer-wise detection and global weighted aggregation, FedLoRa effectively improves the robustness against model poisoning.

\subsubsection{FoundationFL}
FoundationFL\cite{foundationfl} proposes a simple but effective defense framework for FL. The core idea is to enhance existing classical Byzantine-robust aggregation methods (such as Trimmed-mean and Median) by introducing synthetic updates, rather than designing entirely new aggregation rules.
In each global communication round, the server automatically generates synthetic updates that resemble real client updates and aggregates them together using robust aggregation. It works well because the synthetic updates help reduce the variance among all updates, which makes it easier for robust aggregation rules to identify and filter out malicious updates, especially in scenarios with highly heterogeneous (Non-IID) data distributions.

\subsubsection{RFA}
RFA\cite{rfa} is a robust framework for FL designed to address issues caused by malicious attacks while protecting data privacy. Its core approach replaces the weighted average aggregation used in traditional FL with the geometric median. The geometric median minimizes the sum of the Euclidean distances from all client updates to aggregation point, finding a compromise point that remains reliable even when up to 50 \% of the updates are anomalous. The calculation process of geometric median $v^*$ can be expressed as:
\begin{equation}
    v^* = \arg\min_v \sum_{i=1}^m \alpha_i \| v - w_i \|,
\end{equation}
where $m$ is the number of clients and $\alpha_i$ is the aggregation weight of the $i$-th client. RFA relies on the smoothed Weiszfeld algorithm, which iteratively calls a secure average protocol to gradually approximate the geometric median. In each iteration, weights are dynamically adjusted based on the distance between the device updates and the current estimated point, automatically assigning lower weights to anomalous updates.

\subsubsection{RLR}
RLR\cite{rlr} is a defense mechanism that dynamically adjusts the server's learning rate based on the signs of updates. The core of the RLR involves analyzing the signs of client updates to dynamically adjust the server's learning rate. Specifically, the server introduces a learning threshold $\theta$, and for each dimension, it calculates the $i$-th absolute value $S$ of the sum of the update signs at $t$-th global communication round:
\begin{equation}
S_i = \left| \sum_{k \in N_t} \operatorname{sgn}(\Delta_{t,i}^k) \right|,
\end{equation}
where $N_t$ is the number of clients, $\Delta_{t,i}^k$ is the $k$-th client's updates and $\operatorname{sgn}$ is the sign function. if the $S_i$ is greater than or equal to $\theta$, the learning rate remains positive (normal optimization); if it is less than $\theta$, the learning rate becomes negative (reverse optimization, increasing the loss of the malicious task). This can be expressed as:
\begin{equation}
\eta_{i} =
\begin{cases} 
\eta & \text{if } S_i \geq \theta, \\
-\eta & \text{if } S_i < \theta.
\end{cases}
\end{equation}
Finally, the server aggregates the global model based on the adjusted learning rate. In this way, RLR can automatically mitigate the impact of malicious updates, steering the model away from the malicious target and toward the honest target. 

\subsection{Dataset}

\subsubsection{CIFAR10}
The CIFAR10\cite{cifar10} dataset is a popular benchmark dataset used for image classification tasks. It consists of 60,000 32x32 color images in 10 classes, with 6,000 images per class. The classes are: airplane, automobile, bird, cat, deer, dog, frog, horse, ship, and truck. The dataset is commonly used to evaluate machine learning algorithms, particularly in the field of deep learning, for image classification tasks.

\subsubsection{SVHN}
The SVHN\cite{svhn} dataset is a real-world image dataset containing over 600,000 labeled digit images extracted from Google Street View, commonly used for digit recognition and image classification tasks. It includes 32x32 pixel images with digits 0-9, available in both cropped (single-digit) and full-image formats, making it ideal for developing and testing algorithms in real-world, noisy settings.

\subsubsection{CIFAR100}
The CIFAR100\cite{cifar10} dataset is similar in structure to CIFAR10 but is more challenging. It also consists of 60,000 32x32 color images, but these images are divided into 100 fine-grained classes. These 100 classes are further grouped into 20 superclasses, providing a hierarchical structure. For example, the superclass "aquatic mammals" includes classes like beaver, dolphin, and otter. Like CIFAR10, CIFAR100 is split into 50,000 training images and 10,000 test images. The increased number of classes and finer granularity make CIFAR-100 a more complex dataset, often used to evaluate the performance of more advanced models.

\subsection{Default experiment setting}
\label{supp_experiment_setting}

We implement FedGraM and existing defense methods in Python using popular deep learning framework PyTorch. We simulate both cross-device FL and cross-silo FL. In cross-device FL, the FL system includes $500$ clients with $10\%$ of clients are randomly sampled to participate in the training in each communication round. We set model architecture as ResNet 8\cite{resnet} and perform $2000$ rounds of training for cross-device. In cross-silo FL, the FL system includes $50$ clients with all the clients participating in the training in each communication round. We set model architecture as ResNet 18\cite{resnet} and perform $1000$ rounds of training for cross-silo. In both FLs, we set batch size to $32$ and set learning rate to $0.1$. We leverage Stochastic Gradient Descent(SGD) as the optimizer of the training.

\subsection{Demo experiment setting}
\label{supp_demoexpsetting}

In demo experiment, we adopt MinSum attack towards FL system in CIFAR10 with $\beta = 0.2$. We set FedAvg as the aggregation method and record the local models in the 5-th communication round. Separately, we estimate the test accuracy and the Gram matrix norm of each local models. We arrange the clients in descending order based on the corresponding values and show the distribution of norm and accuracy in Figure \ref{introduction}. The Gram matrix is calculated on 10 data samples with one data sample for each class. The test accuracy is calculated on 100 data samples with 10 data samples for each class. The other setting of the experiment follows the default setting.

\section{Additional Experiment Results}

\begin{figure*}[h]
    \centering
    \includegraphics[width=1.0\linewidth]{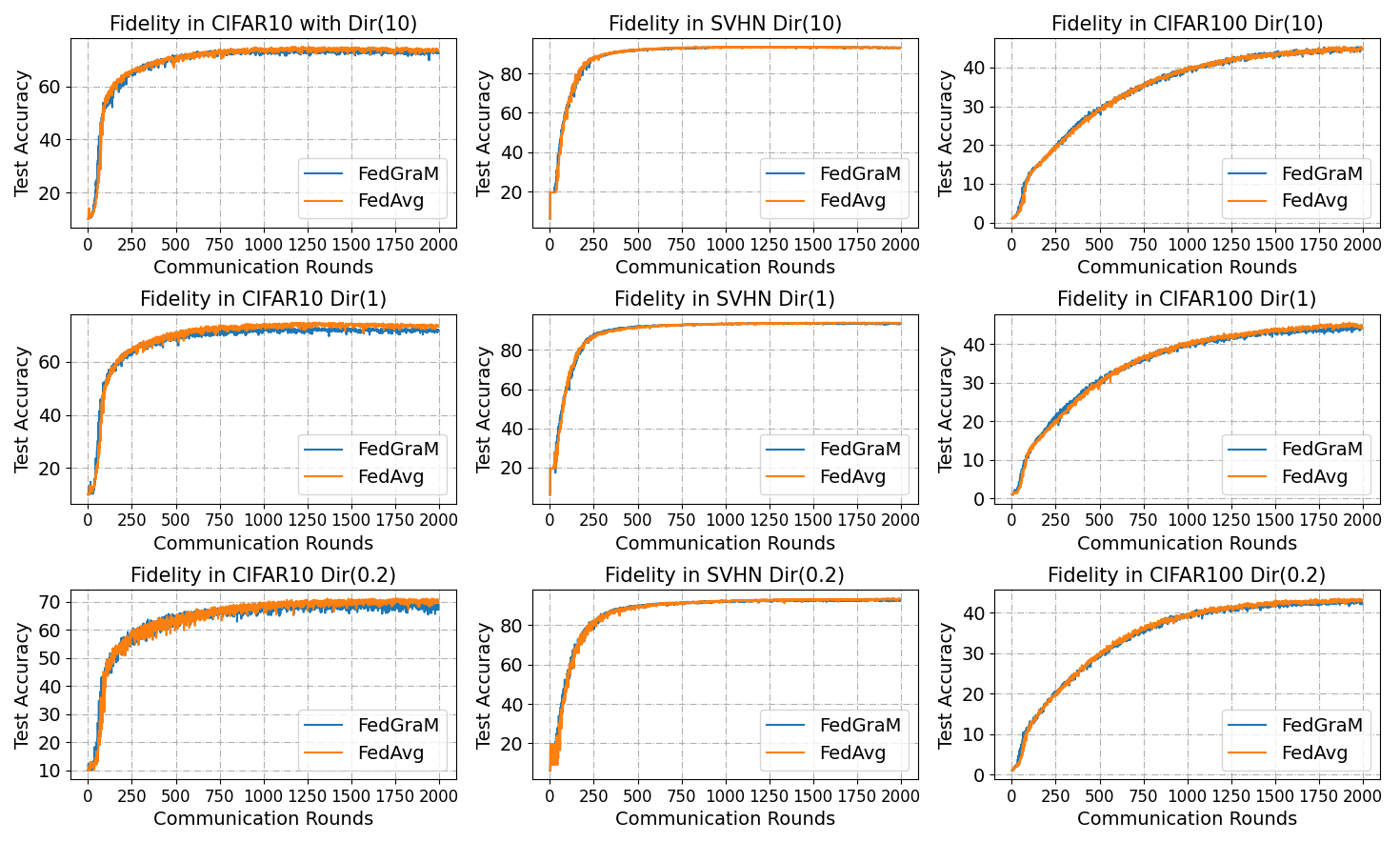}
    \caption{Experiment results of Fidelity of FedGraM. We show the accuracy curves of FedGraM and FedAvg.}
    \label{Fidelity_supp}
\end{figure*}

\subsection{Fidelity of FedGraM}
\label{supp_fidelity}
In this experiment, we aim to evaluate the performance of FedGraM without any attack. As FedGraM removes partial local updates per communication round, it may cause performance sacrifice. Specifically, we estimate the test accuracy of the global model train with FedAvg, and FedGraM to compare our method and standard aggregation method in FL. The experiments are conducted on CIFAR10, SVHN, and CIFAR100 dataset. We set the concentration parameter $\beta$ to $\{10,1,0.2\}$ to simulate the different situations of NonIID. The experiment results are shown in Figure \ref{Fidelity_supp}.

Aligned with the results we showed in the main paper, the FedGraM's performance is similar to the FedAvg's performance in all the situations. Intuitively, FedGraM removes part of the local models in each communication round which may influence the convergence and the generalization of the global model. In fact, such impact is negligible. A potential reason for this phenomenon is that removing partial local models can be treated as a degradation of the client sample rate in each communication round. As the previous study demonstrated, such degradation would only have a tiny influence on the convergence.

\subsection{Impact of C}
\label{supp_impactofc}

We conduct experiments to determine the hyperparameter $C$. We set $C \in \{20\%,30\%,40\%\}$. For NonIID simulation, we set $\beta =1$. We conduct the experiments in CIFAR10, SVHN and CIFAR100. We setup LIE, Fang, MinMax, and MinSum as the untargeted attacks. The results are shown in Table \ref{supp_impactofc_fig}. Accordingly, while $C=40\%$, an excessive number of local models were removed which led to performance degradation in all datasets. FedGraM has shown similar performances with $C=20\%$ and $C=30\%$. However, under certain situations, $C=20\%$ is insufficient to defend against attacks. As a result, setting $C=30\%$ is the safest choice to guarantee robustness. Although there might not be that many malicious clients in the FL system, appropriately removing some local models can mitigate potential threats.

\begin{table}[h]
\centering
\caption{The entire experiment results of the impact of C.}
\setlength{\tabcolsep}{4.1pt}
\renewcommand{\arraystretch}{1.2}
\begin{tabular}{cccccccccc}
\toprule
Dataset & \multicolumn{3}{c}{CIFAR10} & \multicolumn{3}{c}{SVHN}    & \multicolumn{3}{c}{CIFAR100} \\ \hline
C       & 20\%    & 30\%    & 40\%    & 20\%    & 30\%    & 40\%    & 20\%     & 30\%    & 40\%    \\ \bottomrule
LIE     & 71.67\% & 72.47\% & 72.01\% & 91.82\% & 91.27\% & 91.13\% & 43.49\%  & 43.37\% & 43.11\% \\
Fang    & 73.81\% & 73.72\% & 72.80\% & 93.56\% & 93.62\% & 93.06\% & 44.02\%  & 44.65\% & 44.00\% \\
MinMax  & 72.47\% & 72.72\% & 72.53\% & 93.73\% & 93.51\% & 93.61\% & 42.40\%  & 44.02\% & 43.73\% \\
MinSum  & 72.57\% & 71.26\% & 69.45\% & 92.26\% & 93.34\% & 92.95\% & 40.71\%  & 43.59\% & 40.58\% \\ \bottomrule
\end{tabular}
\label{supp_impactofc_fig}
\end{table}

\subsection{Comparison with existing defense methods under different data heterogeneity}
\label{supp_comparison_noniid}
\subsubsection{Cross-device CIFAR10}
We show the comparison results in cross-device scenario in CIFAR10 dataset. Specifically, the performance of evaluated defense methods under LIE, Fang, Label Flip and Dynamic Label Flip are shown in Table \ref{comparison_cifar10_4a}. The performance under MinMax, MinSum, and MPAF attacks are shown in Table \ref{comparison_cifar10_3a}. FedGraM has a good performance in the comparison. Under all kinds of untargeted attacks, it can successfully defend the attacks and maintain the test accuracy of the global model at a high level. Although it is not the best under some situations, it is close to the best which can also demonstrate its effectiveness.
\begin{table}[h]
\centering
\caption{The experiment results of Comparison between FedGraM and existing defense methods in CIFAR10 dataset with cross-device setting under LIE, Fang, Label Flip and Dynamic Label Flip attacks.}
\small
\setlength{\tabcolsep}{3.5pt}
\renewcommand{\arraystretch}{1.2}
\begin{tabular}{ccccccccccccc}
\toprule
Attack       & \multicolumn{3}{c}{LIE}                          & \multicolumn{3}{c}{Fang}                         & \multicolumn{3}{c}{LF}                          & \multicolumn{3}{c}{DLF}                          \\ \hline
$\beta$            & 10             & 1              & 0.2            & 10             & 1              & 0.2            & 10            & 1              & 0.2            & 10             & 1              & 0.2            \\ \bottomrule
FedAvg       & 67.64          & 66.68          & 60.55          & 39.02          & 38.69          & 27.62          & 73.14         & 71.29          & 68.10           & 74.01          & 73.40          & 69.82          \\
Trimean      & 57.64          & 57.31          & 43.77          & 68.34          & 65.98          & 54.77          & 72.74         & 72.37          & 66.47          & 73.10          & 71.72          & 65.46          \\
NormBound    & 67.60          & 66.70          & 60.42          & 69.10           & 68.23          & 62.04          & 72.99         & 71.92          & 68.83          & 74.63          & 73.22          & 69.24          \\
CRFL         & 68.74          & 67.49          & 60.85          & 43.97          & 38.37          & 28.43          & 74.90 & 74.49 & 69.54 & 75.14 & 74.73 & 69.42          \\
FLTrust      & 42.18          & 39.42          & 27.18          & 40.22          & 37.04          & 26.76          & 44.16         & 34.72          & 30.40          & 41.48          & 36.48          & 28.33          \\
FLAME        & 69.32          & 70.01          & 62.55          & 72.25          & 72.20          & 68.69          & 72.23         & 70.76          & 67.90          & 72.22          & 70.54          & 67.61          \\
RONI         & 68.90          & 67.86          & 62.46          & 73.99          & 74.33 & 68.28          & 73.62         & 72.71          & 68.63          & 73.07          & 74.10          & 69.66          \\
Bucket       & 64.76          & 63.93          & 55.59          & 57.32          & 55.05          & 47.30          & 73.02         & 73.39          & 67.65          & 73.97          & 73.20          & 69.72          \\
FedRoLa      & 74.49          & 74.26          & 70.23          & 74.27          & 73.56          & 69.58           & 72.02         & 71.22          & 67.57          & 73.96          & 73.39          & 69.63          \\
MultiKrum    & 63.79          & 63.17          & 54.38          & 72.51          & 71.89          & 66.84          & 71.06         & 71.05          & 65.78          & 72.65          & 71.36          & 68.23          \\
Bulyan       & 42.03          & 39.99          & 30.70          & 68.54          & 66.11          & 49.89          & 67.27         & 65.47          & 47.91          & 67.21          & 65.99          & 50.25          \\
FoundationFL & 66.07          & 64.00          & 54.97          & 72.59          & 71.46          & 62.88          & 72.52         & 71.21          & 64.17          & 72.88          & 72.43          & 63.65          \\
RFA          & 70.53          & 69.69          & 66.26          & 72.17          & 71.20           & 66.72          & 70.56         & 68.43          & 63.51          & 71.26          & 70.54          & 66.71          \\
RLR          &67.70                &67.29                &61.17                &39.06                &37.76                &27.74                &72.05               &71.20                &67.75                &73.63                &72.20                &68.92                \\
FedGraM      & 73.62 & 72.47 & 70.43 & 74.48 & 73.72          & 69.60 & 72.62         & 71.71          & 67.99          & 73.88          & 73.70          & 70.75 \\ \bottomrule
\end{tabular}
\label{comparison_cifar10_4a}
\end{table}

\begin{table}[h]
\centering
\caption{The experiment results of Comparison between FedGraM and existing defense methods in CIFAR10 dataset with cross-device setting under MinMax, MinSum, and MPAF attacks.}
\setlength{\tabcolsep}{6.2pt}
\renewcommand{\arraystretch}{1.2}
\begin{tabular}{cccccccccc}
\toprule
Attack       & \multicolumn{3}{c}{MinMax}                       & \multicolumn{3}{c}{MinSum}                       & \multicolumn{3}{c}{MPAF}                         \\ \hline
$\beta$            & 10             & 1              & 0.2            & 10             & 1              & 0.2            & 10             & 1              & 0.2            \\ \bottomrule
FedAvg       & 49.78          & 48.57          & 39.57          & 23.64          & 18.00             & 17.15          & 29.97          & 28.48          & 25.91          \\
Trimean      & 65.32          & 61.12          & 54.08          & 67.34          & 66.01          & 57.91          & 61.65          & 58.43          & 50.61          \\
NormBound    & 64.41          & 62.92          & 52.31          & 23.12          & 12.36          & 18.48          & 31.44          & 30.24          & 23.45          \\
CRFL         & 59.77          & 46.70           & 38.68          & 26.00            & 20.49          & 19.44          & 29.56          & 29.53          & 26.58          \\
FLTrust      & 24.48          & 19.06          & 17.57          & 41.14          & 36.86          & 27.40           & 32.64            & 32.13             & 30.59             \\
FLAME        & 70.47          & 69.97          & 67.73          & 71.97          & 70.77          & 65.81 & 72.26          & 70.96          & 68.66          \\
RONI         & 65.88          & 62.97          & 50.46          & 22.31          & 23.33          & 18.03          & 36.13          & 33.01          & 30.04          \\
Bucket       & 52.28          & 50.99          & 43.19          & 50.25          & 50.96          & 35.8           & 46.93          & 42.91          & 36.71          \\
FedRoLa      & 75.12          & 73.83          & 70.31          & 74.72          & 74.08          & 70.32          & 65.22          & 63.35          & 57.24          \\
MultiKrum    & 73.38          & 72.60           & 69.70           & 16.99          & 15.34          & 13.49          & 72.96          & 72.20           & 68.91          \\
Bulyan       & 69.25          & 66.93          & 54.47          & 43.51          & 50.38          & 33.51          & 68.37          & 67.24          & 52.98          \\
FoundationFL & 67.75          & 65.12          & 56.75          & 35.65          & 39.48          & 29.87          & 64.49          & 62.12          & 50.45          \\
RFA          & 73.44          & 72.38          & 66.92          & 12.52          & 11.68          & 12.77          & 68.02          & 67.34          & 61.68          \\
RLR          &49.68                &47.14                &40.33                &23.94                &19.96                &15.13                &29.75                & 28.99               &24.51                \\
FedGraM      & 74.49 & 72.72 & 69.46 & 72.25 & 71.26 & 64.90           & 74.24 & 73.41 & 69.56 \\ \bottomrule
\end{tabular}
\label{comparison_cifar10_3a}
\end{table}

\subsubsection{Cross-device SVHN}

We show the comparison results in cross-device scenario in SVHN dataset. Specifically, the performance of evaluated defense methods under LIE, Fang, Label Flip and Dynamic Label Flip are shown in Table \ref{comparison_svhn_4a}. The performance under MinMax, MinSum, and MPAF attacks are shown in Table \ref{comparison_svhn_3a}. Accordingly, FedGraM's performance in SVHN is better than its performance in CIFAR10 as it achieve the best accuracy in more situations. SVHN is an easier classification task compared with CIFAR10 which further facilitate the robustness of FedGraM.

\begin{table}[h]
\centering
\caption{The experiment results of Comparison between FedGraM and existing defense methods in SVHN dataset with cross-device setting under LIE, Fang, Label Flip and Dynamic Label Flip attacks.}
\small
\setlength{\tabcolsep}{3.5pt}
\renewcommand{\arraystretch}{1.2}
\begin{tabular}{ccccccccccccc}
\toprule
Attack       & \multicolumn{3}{c}{LIE}                          & \multicolumn{3}{c}{Fang}                         & \multicolumn{3}{c}{LF}                           & \multicolumn{3}{c}{DLF}                          \\ \hline
$\beta$            & 10             & 1              & 0.2            & 10             & 1              & 0.2            & 10             & 1              & 0.2            & 10             & 1              & 0.2            \\ \bottomrule
FedAvg       & 91.41          & 91.22          & 89.22          & 19.63          & 19.69          & 19.62          & 92.65          & 92.59          & 91.27          & 92.93          & 93.02          & 92.68          \\
Trimean      & 87.84          & 86.37          & 75.62          & 92.37          & 91.51          & 84.23          & 92.39          & 92.48          & 91.81          & 92.54          & 92.68          & 91.62          \\
NormBound    & 90.70           & 90.64          & 88.49          & 91.76          & 91.71          & 90.27          & 92.15          & 92.03          & 91.92          & 93.31          & 93.23          & 92.66          \\
CRFL         & 91.81          & 91.84 & 90.31          & 19.72          & 19.61          & 19.71          & 93.26 & 93.33 & 92.92 & 93.63 & 93.48 & 92.77          \\
FLTrust      & 63.37          & 45.98          & 34.20           & 62.54          & 45.97          & 20.01          & 50.29          & 50.31          & 23.4           & 26.95          & 48.9           & 26.99          \\
FLAME        & 91.44          & 91.98          & 88.98          & 93.13          & 93.26          & 92.70           & 93.03          & 93.09          & 92.72          & 93.05          & 93.14          & 92.80           \\
RONI         & 91.73          & 91.55          & 89.76          & 19.66          & 21.75 & 19.65          & 93.41          & 93.29          & 92.20           & 93.7           & 93.74 & 93.57 \\
Bucket       & 90.26          & 89.73          & 86.22          & 85.58          & 85.20           & 67.97          & 93.21          & 93.47 & 92.43          & 93.57          & 93.36          & 92.90           \\
FedRoLa      & 93.33          & 93.53          & 93.16          & 93.68          & 93.30           & 92.90           & 93.56          & 93.31          & 92.42          & 93.36          & 93.43          & 93.02          \\
MultiKrum    & 90.48          & 89.08          & 87.50           & 93.16          & 93.18          & 92.30           & 92.98          & 92.76          & 91.71          & 93.25          & 93.60           & 92.92          \\
Bulyan       & 62.35          & 58.72          & 24.08          & 91.87          & 90.99          & 86.07          & 91.31          & 91.17          & 84.51          & 91.64          & 90.78          & 86.81          \\
FoundationFL & 91.32          & 90.43          & 86.55          & 92.82          & 92.28          & 90.22          & 92.87          & 92.58          & 91.36          & 93.41          & 93.26          & 90.85          \\
RFA          & 92.71          & 92.73          & 91.93          & 92.56          & 92.64          & 92.27          & 92.38          & 92.51          & 91.03          & 93.22          & 93.29          & 92.23          \\
RLR          &91.31                & 91.26               & 89.62               &19.64                & 19.69               & 19.77               &91.98                &91.57                & 90.88               &92.21                & 91.87               & 90.52               \\
FedGraM      & 92.12 & 91.27 & 91.35 & 93.67 & 93.62 & 93.23 & 93.57 & 93.46          & 93.22 & 93.72 & 93.51          & 92.90 \\ \bottomrule
\end{tabular}
\label{comparison_svhn_4a}
\end{table}

\begin{table}[h]
\centering
\caption{The experiment results of Comparison between FedGraM and existing defense methods in SVHN dataset with cross-device setting under MinMax, MinSum, and MPAF attacks.}
\setlength{\tabcolsep}{6.2pt}
\renewcommand{\arraystretch}{1.2}
\begin{tabular}{cccccccccc}
\toprule
Attack       & \multicolumn{3}{c}{MinMax}                       & \multicolumn{3}{c}{MinSum}                       & \multicolumn{3}{c}{MPAF}                        \\ \hline
$\beta$            & 10             & 1              & 0.2            & 10             & 1              & 0.2            & 10             & 1             & 0.2            \\ \bottomrule
FedAvg       & 79.34          & 66.85          & 43.09          & 19.76          & 20.1           & 19.60           & 21.62          & 20.64         & 19.59          \\
Trimean      & 89.62          & 87.07          & 79.61          & 92.19          & 91.21          & 87.10           & 88.36          & 87.37         & 79.42          \\
NormBound    & 89.99          & 88.33          & 81.72          & 19.58          & 23.27          & 20.20           & 19.97          & 20.51         & 19.59          \\
CRFL         & 85.03          & 76.17          & 43.31          & 19.76          & 19.92          & 19.84          & 20.26          & 20.24         & 19.59          \\
FLTrust      & 19.59          & 19.58          & 19.58          & 50.45          & 36.37          & 42.17          & 6.69           & 6.69          & 6.69           \\
FLAME        & 92.73          & 92.70           & 92.50           & 93.05          & 93.47 & 91.54 & 93.06          & 93.29         & 92.63          \\
RONI         & 92.99          & 91.23          & 34.36          & 21.29          & 19.60           & 19.84          & 23.50           & 21.85         & 19.61          \\
Bucket       & 84.43          & 74.74          & 55.33          & 84.55          & 56.19          & 20.49          & 63.62          & 59.13         & 33.17          \\
FedRoLa      & 93.41          & 93.41          & 93.24          & 93.61          & 94.01          & 93.36          & 89.65          & 88.36         & 82.95          \\
MultiKrum    & 93.12          & 92.96          & 92.98          & 19.58          & 19.58          & 19.58          & 93.12          & 93.00         & 92.67          \\
Bulyan       & 91.93          & 91.27          & 86.77          & 66.21          & 73.85          & 22.88          & 91.67          & 91.44         & 84.35          \\
FoundationFL & 91.02          & 90.17          & 83.04          & 22.70           & 19.77          & 19.59          & 90.85          & 90.09         & 86.45          \\
RFA          & 92.65          & 92.62          & 91.55          & 19.59          & 19.58          & 19.58          & 90.44          & 89.16         & 87.31          \\
RLR          & 80.33               &63.89                &47.32                &  19.58              &  19.58              &   19.57            &20.07                &20.21               & 19.64               \\
FedGraM      & 93.88 & 93.51 & 93.08 & 93.80 & 93.34 & 92.67          & 93.26 & 93.50 & 93.26 \\ \bottomrule
\end{tabular}
\label{comparison_svhn_3a}
\end{table}

\subsubsection{Cross-device CIFAR100}

We show the comparison results in cross-device scenario in SVHN dataset. Specifically, the performance of evaluated defense methods under LIE, Fang, Label Flip and Dynamic Label Flip are shown in Table \ref{comparison_cifar100_4a}. The performance under MinMax, MinSum, and MPAF attacks are shown in Table \ref{comparison_cifar100_3a}. CIFAR100 classification is the most difficult task among all the evaluated three tasks. As shown in our results, many defense methods fall short in defending in CIFAR100. However, our method FedGraM is still effectiveness. Only few methods can be effective as FedGraM in defending all kinds of attacks under all the NonIID situations.

\begin{table}[h]
\centering
\caption{The experiment results of Comparison between FedGraM and existing defense methods in CIFAR100 dataset with cross-device setting under LIE, Fang, Label Flip and Dynamic Label Flip attacks.}
\small
\setlength{\tabcolsep}{3.5pt}
\renewcommand{\arraystretch}{1.2}
\begin{tabular}{ccccccccccccc}
\toprule
Attack       & \multicolumn{3}{c}{LIE}                          & \multicolumn{3}{c}{Fang}                         & \multicolumn{3}{c}{LF}                           & \multicolumn{3}{c}{DLF}                          \\ \hline
$\beta$            & 10             & 1              & 0.2            & 10             & 1              & 0.2            & 10             & 1              & 0.2            & 10             & 1              & 0.2            \\ \bottomrule
FedAvg       & 33.93          & 33.68          & 32.66          & 6.19           & 6.04           & 2.43           & 44.77          & 44.04          & 43.48          & 45.03          & 43.52          & 43.14          \\
Trimean      & 20.01          & 17.77          & 9.56           & 30.45          & 25.66          & 8.39           & 43.00             & 40.89          & 25.95          & 42.04          & 40.09          & 26.28          \\
NormBound    & 33.16          & 32.81          & 33.10           & 6.11           & 4.05           & 1.87           & 43.85          & 43.54          & 42.07          & 45.11          & 43.18          & 41.22          \\
CRFL         & 33.79          & 34.46          & 32.73          & 5.93           & 5.48           & 2.32           & 44.72 & 44.38 & 42.99 & 43.23 & 43.40  & 43.10  \\
FLTrust      & 7.25           & 6.27           & 5.53           & 9.33           & 7.00              & 6.97           & 6.17           & 6.22           & 6.08           & 5.24           & 7.08           & 6.08           \\
FLAME        & 33.56          & 36.38          & 33.33          & 42.40           & 42.77          & 40.80           & 42.72          & 42.14          & 41.34          & 43.08          & 41.78          & 40.63          \\
RONI         & 35.02          & 35.88          & 31.6           & 43.92          & 44.29 & 41.63          & 43.59          & 43.76          & 42.29          & 43.96          & 43.42          & 43.50           \\
Bucket       & 27.95          & 27.03          & 20.48          & 14.15          & 13.53          & 7.46           & 42.48          & 43.86          & 39.33          & 43.24          & 43.79          & 36.96          \\
FedRoLa      & 45.30           & 44.69          & 43.71          & 45.49          & 45.18          & 43.33          & 42.79          & 42.98          & 40.93          & 44.31          & 43.95          & 42.40           \\
MultiKrum    & 28.93          & 26.69          & 25.83          & 42.08          & 41.13          & 40.47          & 40.37          & 40.49          & 39.83          & 41.66          & 41.87          & 40.22          \\
Bulyan       & 9.92           & 7.29           & 3.79           & 31.16          & 27.25          & 11.41          & 30.36          & 25.35          & 9.67           & 30.72          & 25.33          & 10.54          \\
FoundationFL & 28.07          & 27.98          & 21.77          & 34.87          & 32.78          & 27.11          & 37.59          & 37.72          & 31.74          & 37.66          & 36.39          & 28.66          \\
RFA          & 38.12          & 38.18          & 37.12          & 38.74          & 38.69          & 37.44          & 38.16          & 37.86          & 36.72          & 39.40           & 39.19          & 37.31          \\
RLR          &33.79                &33.19                &32.82                &6.94                &7.04                &2.21                &43.68                &42.77                &42.12                &43.60               &43.35                & 42.16               \\
FedGraM      & 43.46 & 43.37 & 42.60 & 45.26 & 44.65 & 43.67 & 44.88 & 43.70           & 43.32 & 45.15 & 43.93 & 42.87 \\ \bottomrule
\end{tabular}
\label{comparison_cifar100_4a}
\end{table}

\begin{table}[h]
\centering
\caption{The experiment results of Comparison between FedGraM and existing defense methods in CIFAR100 dataset with cross-device setting under MinMax, MinSum, and MPAF attacks.}
\setlength{\tabcolsep}{6.2pt}
\renewcommand{\arraystretch}{1.2}
\begin{tabular}{cccccccccc}
\toprule
Attack       & \multicolumn{3}{c}{MinMax}                       & \multicolumn{3}{c}{MinSum}                      & \multicolumn{3}{c}{MPAF}                        \\ \hline
$\beta$            & 10             & 1              & 0.2            & 10            & 1              & 0.2            & 10             & 1              & 0.2           \\ \bottomrule
FedAvg       & 14.71          & 15.95          & 13.9           & 1.77          & 1.99           & 2.18           & 3.73           & 3.77           & 4.18          \\
Trimean      & 23.82          & 23.71          & 15.97          & 31.15         & 25.52          & 9.15           & 23.23          & 24.49          & 15.73         \\
NormBound    & 13.7           & 13.07          & 12.42          & 2.25          & 1.91           & 1.80            & 3.95           & 3.93           & 3.98          \\
CRFL         & 7.69           & 9.21           & 7.00              & 1.98          & 1.66           & 1.77           & 3.52           & 4.04           & 3.95          \\
FLTrust      & 2.56           & 3.03           & 2.33           & 5.92          & 6.62           & 5.56           & 5.89           & 5.47           & 5.39          \\
FLAME        & 24.74          & 29.55          & 26.41          & 39.17         & 40.84          & \textbf{37.37} & 43.08          & 42.41          & 41.16         \\
RONI         & 33.94          & 31.66          & 32.32          & 4.45          & 3.54           & 2.89           & 9.53           & 7.63           & 7.59          \\
Bucket       & 16.29          & 17.42          & 16.79          & 8.99          & 9.15           & 5.23           & 9.32           & 9.78           & 8.85          \\
FedRoLa      & 46.13          & 45.63          & 43.94          & 44.19         & 45.3           & 43.63          & 20.27          & 21.85          & 20.93         \\
MultiKrum    & 42.77          & 41.86          & 40.56          & 1.61          & 1.69           & 1.51           & 42.29          & 40.97          & 39.97         \\
Bulyan       & 33.27          & 28.34          & 11.18          & 9.63          & 6.8            & 3.75           & 32.29          & 27.41          & 10.39         \\
FoundationFL & 28.61          & 30.14          & 24.09          & 3.51          & 5.85           & 6.86           & 26.65          & 26.81          & 20.53         \\
RFA          & 42.19          & 41.45          & 39.85          & 1.78          & 1.35           & 1.24           & 24.77          & 26.32          & 23.53         \\
RLR          &15.45                &16.25                &15.05                &1.86               &1.67                &1.98                &4.20                &3.72                & 3.96               \\
FedGraM      & 44.88 & 44.02 & 43.81 & 42.90 & 43.59 & 37.76 & 45.23 & 44.85 & 43.70 \\ \bottomrule
\end{tabular}
\label{comparison_cifar100_3a}
\end{table}

\subsubsection{Cross-silo}
\label{supp_comparison_silo}

\begin{table}[h]
\centering
\caption{The experiment results of Comparison between FedGraM and existing defense methods in CIFAR10 dataset in cross-silo scenario}
\renewcommand{\arraystretch}{1.2}
\begin{tabular}{ccccccccc}
\toprule
\multicolumn{9}{c}{CIFAR10 (Cross-Silo)}                                                          \\ \hline
Attack                  & $\beta$ & FedAvg & Trimean & NormBound & CRFL  & FLTrust & RONI  & FedGraM \\ \bottomrule
\multirow{3}{*}{LIE}    & 10   & 80.42  & 77.30   & 78.67     & 79.20 & 78.89   & 79.06 & 79.21   \\
                        & 1    & 79.23  & 74.60   & 77.37     & 77.71 & 77.42   & 77.75 & 78.35   \\
                        & 0.2  & 71.07  & 50.08   & 71.13     & 71.32 & 72.39   & 71.92 & 71.63   \\
\multirow{3}{*}{Fang}   & 10   & 65.17  & 76.20   & 50.10     & 43.82 & 63.08   & 64.72 & 78.74   \\
                        & 1    & 58.44  & 74.26   & 38.40     & 38.75 & 74.93   & 80.42 & 78.59   \\
                        & 0.2  & 32.99  & 60.05   & 23.69     & 22.53 & 67.35   & 65.27 & 67.86   \\
\multirow{3}{*}{MinMax} & 10   & 72.32  & 77.93   & 76.89     & 68.11 & 79.72   & 57.00 & 79.04   \\
                        & 1    & 65.59  & 75.29   & 67.70     & 66.34 & 78.30   & 49.52 & 78.26   \\
                        & 0.2  & 56.68  & 61.59   & 45.46     & 47.88 & 70.28   & 37.71 & 69.53   \\
\multirow{3}{*}{MinSum} & 10   & 13.06  & 77.15   & 28.67     & 82.97 & 17.03   & 25.56 & 78.74   \\
                        & 1    & 15.70  & 73.35   & 17.29     & 77.90 & 16.63   & 23.45 & 76.27   \\
                        & 0.2  & 13.80   & 61.65   & 13.30     & 57.58 & 14.80   & 18.75 & 75.39   \\ \bottomrule
\end{tabular}
\label{comparison_crosssilo_cifar10}
\end{table}

\begin{table}[h]
\centering
\caption{The experiment results of Comparison between FedGraM and existing defense methods in SVHN dataset in cross-silo scenario}
\renewcommand{\arraystretch}{1.2}
\begin{tabular}{ccccccccc}
\toprule
\multicolumn{9}{c}{SVHN (Cross-Silo)}                                                             \\ \hline
Attack                  & $\beta$ & FedAvg & Trimean & NormBound & CRFL  & FLTrust & RONI  & FedGraM \\ \bottomrule
\multirow{3}{*}{LIE}    & 10   & 94.88  & 93.53   & 94.19     & 94.18 & 94.21   & 94.44 & 93.34   \\
                        & 1    & 94.58  & 93.21   & 94.16     & 94.46 & 94.41   & 94.33 & 93.85   \\
                        & 0.2  & 93.72  & 90.02   & 93.66     & 93.30 & 19.64   & 92.55 & 92.79   \\
\multirow{3}{*}{Fang}   & 10   & 19.59  & 93.60   & 19.59     & 19.66 & 89.87   & 19.74 & 94.34   \\
                        & 1    & 87.43  & 93.99   & 85.45     & 73.24 & 91.37   & 19.79 & 94.45   \\
                        & 0.2  & 19.12  & 90.28   & 9.69      & 10.59 & 92.44   & 19.54 & 93.81   \\
\multirow{3}{*}{MinMax} & 10   & 91.00  & 93.41   & 91.19     & 88.99 & 87.64   & 87.50 & 94.32   \\
                        & 1    & 84.06  & 92.88   & 89.63     & 80.52 & 91.13   & 86.57 & 94.29   \\
                        & 0.2  & 76.08  & 8614    & 80.27     & 64.02 & 89.23   & 86.11 & 92.71    \\
\multirow{3}{*}{MinSum} & 10   & 20.68  & 94.03   & 95.10     & 95.52 & 19.58   & 19.58 & 93.81   \\
                        & 1    & 19.58  & 93.42   & 98.63     & 22.27 & 19.71   & 19.62 & 92.54   \\
                        & 0.2  & 19.61  & 91.49   & 26.51     & 94.63 & 19.65   & 19.47 & 89.66   \\ \bottomrule
\end{tabular}
\label{comparison_crosssilo_svhn}
\end{table}

We also evaluate the performance of FedGraM in cross-silo scenario. To distinguish from cross-device scenario, we make two main modifications of the experiment setting. Firs,t we change the model architecture from ResNet8\cite{resnet} to ResNet18\cite{resnet} as in cross-silo scenarios, the client is expected to equip a better hardware device to perform local training. Second, all the clients will be selected to participate in the training in each communication round as the client should have better equipment to guarantee communication with the server. The experiments are conducted in CIFAR10 dataset and SVHN dataset. We set $\beta \in \{10,1,0.2\}$. Besides FedGraM, we implement other 6 methods including FedAvg, Trimean, NormBound, CRFL, FLTrust, and RONI. We implement LIE attack, Fang attack, MinMax attack, and MinSum attack to evaluate the robustness. The experiments are shown in Table \ref{comparison_crosssilo_svhn} and Table \ref{comparison_crosssilo_cifar10}. Accordingly, some existing defense methods have shown their effectiveness in cross-silo scenarios. Especially for FLTrust, it is weak to defend against any attack in the cross-device scenarios but robust to defend LIE, Fang and MinMax attack in the cross-silo scenario. A potential reason is that in cross-device scenarios, the local models of the clients are intermittently participating in the communication round whereas the root model is trained in each communication round. The training of local models and the root model is inconsistent, leading to that the root model can not serve as the standard of local models. In cross-silo scenarios, both the local models and the root model participate in each communication round and the root model can be a good standard for local models. Regarding the performance of FedGraM, its robustness is retained and the performance of FedGraM is still ranked as the highest level.

\subsection{Comparison with existing defense methods under different malicious client ratio}
\label{supp_comparison_attackraio}

\begin{table}[h]
\centering
\caption{The experiment results of the impact of malicious clients ratio. We record the highest accuracy($\%$) achieved by the global model during the training to reflect the performance of each defense method. }
\small
\setlength{\tabcolsep}{4.1pt}
\renewcommand{\arraystretch}{1.2}
\begin{tabular}{ccccccccccccc}
\toprule
\multicolumn{13}{c}{CIFAR10}                                                                                             \\ \bottomrule
Attacks   & \multicolumn{3}{c}{LIE} & \multicolumn{3}{c}{Fang} & \multicolumn{3}{c}{MinMax} & \multicolumn{3}{c}{MinSum} \\
Ratio     & 15\%   & 10\%   & 5\%   & 15\%   & 10\%   & 5\%    & 15\%    & 10\%    & 5\%    & 15\%    & 10\%    & 5\%    \\ \bottomrule
FedAvg    & 59.74  & 66.68  & 72.58 & 29.52  & 39.68  & 68.40  & 52.38   & 48.57   & 62.02  & 18.08   & 18.00   & 30.69  \\
Trimean   & 48.76  & 57.31  & 65.61 & 57.54  & 65.98  & 70.40  & 50.77   & 61.12   & 70.74  & 58.25   & 66.01   & 69.98  \\
NormBound & 60.45  & 66.70  & 69.93 & 48.99  & 68.23  & 71.09  & 51.68   & 62.92   & 71.50  & 17.46   & 12.36   & 33.25  \\
CRFL      & 62.65  & 67.49  & 71.56 & 28.19  & 38.37  & 67.59  & 32.67   & 46.70   & 68.09  & 15.51   & 20.49   & 47.38  \\
FLTrust   & 38.76  & 39.42  & 37.14 & 39.17  & 37.04  & 32.90  & 18.09   & 19.06   & 20.32  & 37.89   & 36.86   & 36.17  \\
RONI      & 62.37  & 67.87  & 72.84 & 70.36  & 74.40  & 74.21  & 48.52   & 64.67   & 70.61  & 19.6    & 25.07   & 69.03  \\
FedGraM   & 72.57  & 72.47  & 73.91 & 73.94  & 73.72  & 74.16  & 73.06   & 72.72   & 72.98  & 63.26   & 71.26   & 73.29  \\ \toprule
\multicolumn{13}{c}{SVHN}                                                                                                \\ \bottomrule
Attack    & \multicolumn{3}{c}{LIE} & \multicolumn{3}{c}{Fang} & \multicolumn{3}{c}{MinMax} & \multicolumn{3}{c}{MinSum} \\
Ratio     & 15\%   & 10\%   & 5\%   & 15\%   & 10\%   & 5\%    & 15\%    & 10\%    & 5\%    & 15\%    & 10\%    & 5\%    \\ \bottomrule
FedAvg    & 89.32  & 91.22  & 93.13 & 19.58  & 19.69  & 91.18  & 78.71   & 66.85   & 87.34  & 19.58   & 20.10   & 23.89  \\
Trimean   & 82.41  & 86.37  & 90.89 & 86.61  & 91.54  & 92.70  & 76.54   & 87.07   & 91.54  & 87.27   & 91.21   & 92.66  \\
NormBound & 88.44  & 90.64  & 92.77 & 74.37  & 91.71  & 93.02  & 83.83   & 91.62   & 91.62  & 19.59   & 23.27   & 22.03  \\
CRFL      & 19.58  & 91.84  & 93.84 & 19.58  & 19.61  & 91.44  & 19.58   & 76.17   & 91.01  & 19.61   & 19.92   & 20.03  \\
FLTrust   & 36.55  & 45.98  & 44.88 & 56.00  & 45.97  & 52.72  & 19.58   & 19.58   & 19.58  & 41.04   & 36.37   & 48.67  \\
RONI      & 91.21  & 91.55  & 93.22 & 19.64  & 21.75  & 93.20  & 36.87   & 91.23   & 93.15  & 19.59   & 19.60   & 20.99  \\
FedGraM   & 93.26  & 91.27  & 92.30 & 84.64  & 93.62  & 93.25  & 93.56   & 93.51   & 93.66  & 85.77   & 93.34   & 93.74  \\ \bottomrule
\end{tabular}
\label{Attackratio_supp}
\end{table}

We show the entire experiment results for the impact of malicious clients ratio. We implement Trimean, Norm Bound, CRFL, FLTrust, RONI, and FedAvg. We applied 4 untargeted attacks including LIE, Fang, MinMax, and MinSum. The experiments are conducted in CIFAR10 dataset, SVHN dataset, and CIFAR100 dataset. We set $\beta = 1$. The results in CIFAR10 dataset, and SVHN dataset are shown in Table \ref{Attackratio_supp}. The experiment results are aligned with the results we show in the main paper. Accordingly, the performance of FedGraM is consistent with the ratio of the malicious clients. For most situations, FedGraM achieves the best performance among evaluated defense methods. More importantly, it is obvious that the performance of other defense methods degrades as the ratio of malicious clients increases. On the contrary, the performance of FedGraM is not impacted by the ratio of malicious clients. However, in some situations, FedGraM is worse than other defense methods. In most of these situations, the FedGraM achieves a similar performance as the best method which can also demonstrate its effectiveness. In SVHN dataset while $15\%$ of clients are malicious clients who launch Fang attack on the FL system, the accuracy of FedGraM is obviously lower than Trimean's accuracy. We treat this as the drawback of FedGraM as it does perform well in some specific scenarios. We hope to improve its performance in the future work.

\subsection{Performance Against Backdoor Attack}

\begin{table}[h]
\centering
\caption{The experiment results for performance against backdoor attacks. Different from the previous experiment, in this experiment, we record the main task accuracy and the backdoor accuracy of the final model.}
\setlength{\tabcolsep}{7pt}
\renewcommand{\arraystretch}{1.2}
\begin{tabular}{ccccccc}
\toprule
\multirow{2}{*}{Dataset}  & \multirow{2}{*}{$\beta$} & \multirow{2}{*}{Attack} & \multicolumn{2}{c}{Trimean} & \multicolumn{2}{c}{FedGraM} \\
                          &                       &                         & Main Acc   & Backdoor Acc   & Main Acc   & Backdoor Acc   \\ \bottomrule
\multirow{6}{*}{CIFAR10}  & \multirow{2}{*}{10}   & Scaling                 & 74.86      & 3.662.53       & 75.49      & 2.20           \\
                          &                       & DBA                     & 73.83      & 2.74           & 75.22      & 2.28           \\
                          & \multirow{2}{*}{1}    & Scaling                 & 73.75      & 3.66           & 73.56         & 2.82           \\
                          &                       & DBA                     & 71.89      & 4.38           & 74.40      & 2.47           \\
                          & \multirow{2}{*}{0.2}  & Scaling                 & 67.00      & 2.80           & 70.44      & 3.00           \\
                          &                       & DBA                     & 61.51      & 9.15           & 70.03      & 4.42           \\
\multirow{6}{*}{SVHN}     & \multirow{2}{*}{10}   & Scaling                 & 93.46      & 0.42           & 93.52      & 0.49           \\
                          &                       & DBA                     & 92.87      & 1.20           & 93.82      & 0.40           \\
                          & \multirow{2}{*}{1}    & Scaling                 & 93.46      & 0.78           & 93.83      & 0.40           \\
                          &                       & DBA                     & 93.27      & 0.72           & 93.84      & 0.60           \\
                          & \multirow{2}{*}{0.2}  & Scaling                 & 93.47      & 0.81           & 93.37      & 0.56           \\
                          &                       & DBA                     & 91.08      & 1.71           & 93.19      & 0.54           \\
\multirow{6}{*}{CIFAR100} & \multirow{2}{*}{10}   & Scaling                 & 43.56      & 0.28           & 45.22      & 0.14           \\
                          &                       & DBA                     & 42.70      & 0.31           & 45.11      & 0.38           \\
                          & \multirow{2}{*}{1}    & Scaling                 & 40.88      & 0.44           & 45.10      & 0.17           \\
                          &                       & DBA                     & 38.94      & 0.26           & 43.75      & 0.70           \\
                          & \multirow{2}{*}{0.2}  & Scaling                 & 22.59      & 1.57           & 44.20      & 0.52           \\
                          &                       & DBA                     & 20.89      & 3.64           & 43.70      & 0.28           \\ \bottomrule
\end{tabular}
\end{table}

We also conduct experiments to evaluate the performance of FedGraM against backdoor attacks. It is worth noting that FedGraM is not designed to defend against backdoor attacks. We aim to evaluate its potential in defending the back door attacks. Specifically, we implement two backdoor attacks including Scaling and DBA\cite{dba}. We compare the performance of FedGraM with Trimean and NormBound. The experiments are conducted in CIFAR10, SVHN, and CIFAR100. We set $\beta \in \{ 10,1,0.2\}$.  The experiment results are shown in Table \ref{backdoor}. We record the main task accuracy and the backdoor accuracy of the final model. The main task accuracy is the accuracy estimated on the regular test set. The backdoor accuracy is estimated on the data with triggers. Accordingly, FedGraM has shown its basic ability to defend against backdoor attacks in comparison with NormBound and Trimean. A potential reason for its robustness against the backdoor is that both DBA and Scaling attack demands to scale the local models to make sure the backdoor can be successfully injected into the global model. Such scaling may affect the embedding space of the local models which can be captured by FedGraM. As a result, FedGraM can detect malicious models with a backdoor injected. In this experiment, we only test the performance of FedGraM under simple backdoor attacks. We are exploring extending this backdoor robustness to more powerful attacks in our future work.


\subsection{Computation overhead on server}

We simply evaluate the computation overhead of FedGraM to the server. We record the extra computation overhead of FedGraM compared with FedAvg. We run our experiment on the workstation with Intel(R) Xeon(R) Gold 6226R CPU and NVIDIA A100 GPU. For CIFAR10 classification, the average extra time consumption for FedGraM is 692.49 ms and the average memory consumption is 30MB. 

\section{Impact Statement}

The proposed method FedGraM is an robust aggregation method in Federated Learning which supposed to enhance the robustness of FL against malicious clients. We believe FedGraM has positive impact to the society that it can be utilized in a wide range research filed to guarantee the robustness.


\newpage
\section*{NeurIPS Paper Checklist}

The checklist is designed to encourage best practices for responsible machine learning research, addressing issues of reproducibility, transparency, research ethics, and societal impact. Do not remove the checklist: {\bf The papers not including the checklist will be desk rejected.} The checklist should follow the references and follow the (optional) supplemental material.  The checklist does NOT count towards the page
limit. 

Please read the checklist guidelines carefully for information on how to answer these questions. For each question in the checklist:
\begin{itemize}
    \item You should answer \answerYes{}, \answerNo{}, or \answerNA{}.
    \item \answerNA{} means either that the question is Not Applicable for that particular paper or the relevant information is Not Available.
    \item Please provide a short (1–2 sentence) justification right after your answer (even for NA). 
\end{itemize}

{\bf The checklist answers are an integral part of your paper submission.} They are visible to the reviewers, area chairs, senior area chairs, and ethics reviewers. You will be asked to also include it (after eventual revisions) with the final version of your paper, and its final version will be published with the paper.

The reviewers of your paper will be asked to use the checklist as one of the factors in their evaluation. While "\answerYes{}" is generally preferable to "\answerNo{}", it is perfectly acceptable to answer "\answerNo{}" provided a proper justification is given (e.g., "error bars are not reported because it would be too computationally expensive" or "we were unable to find the license for the dataset we used"). In general, answering "\answerNo{}" or "\answerNA{}" is not grounds for rejection. While the questions are phrased in a binary way, we acknowledge that the true answer is often more nuanced, so please just use your best judgment and write a justification to elaborate. All supporting evidence can appear either in the main paper or the supplemental material, provided in appendix. If you answer \answerYes{} to a question, in the justification please point to the section(s) where related material for the question can be found.

IMPORTANT, please:
\begin{itemize}
    \item {\bf Delete this instruction block, but keep the section heading ``NeurIPS Paper Checklist"},
    \item  {\bf Keep the checklist subsection headings, questions/answers and guidelines below.}
    \item {\bf Do not modify the questions and only use the provided macros for your answers}.
\end{itemize}


\begin{enumerate}

\item {\bf Claims}
    \item[] Question: Do the main claims made in the abstract and introduction accurately reflect the paper's contributions and scope?
    \item[] Answer: \answerYes{} 
    \item[] Justification: We have reflected the main claims in our methodology and empirical evaluations.

\item {\bf Limitations}
    \item[] Question: Does the paper discuss the limitations of the work performed by the authors?
    \item[] Answer: \answerYes{} 
    \item[] Justification: We have discussed the limitation of the paper in Section 6.

\item {\bf Theory assumptions and proofs}
    \item[] Question: For each theoretical result, does the paper provide the full set of assumptions and a complete (and correct) proof?
    \item[] Answer: \answerNA{} 
    \item[] Justification: The paper does not include theoretical results.

    \item {\bf Experimental result reproducibility}
    \item[] Question: Does the paper fully disclose all the information needed to reproduce the main experimental results of the paper to the extent that it affects the main claims and/or conclusions of the paper (regardless of whether the code and data are provided or not)?
    \item[] Answer: \answerYes{} 
    \item[] Justification: The paper has clearly stated the methodology for reproducibility.

\item {\bf Open access to data and code}
    \item[] Question: Does the paper provide open access to the data and code, with sufficient instructions to faithfully reproduce the main experimental results, as described in supplemental material?
    \item[] Answer: \answerNo{} 
    \item[] Justification: The paper does not provide data and code.

\item {\bf Experimental setting/details}
    \item[] Question: Does the paper specify all the training and test details (e.g., data splits, hyperparameters, how they were chosen, type of optimizer, etc.) necessary to understand the results?
    \item[] Answer: \answerYes{} 
    \item[] Justification: The paper has provided the detailed setting of all the experiments.

\item {\bf Experiment statistical significance}
    \item[] Question: Does the paper report error bars suitably and correctly defined or other appropriate information about the statistical significance of the experiments?
    \item[] Answer: \answerNo{} 
    \item[] Justification: The paper does not provide error bars.

\item {\bf Experiments compute resources}
    \item[] Question: For each experiment, does the paper provide sufficient information on the computer resources (type of compute workers, memory, time of execution) needed to reproduce the experiments?
    \item[] Answer: \answerYes{} 
    \item[] Justification: The paper has provided information on the computer resources and the hardware environment.
    
\item {\bf Code of ethics}
    \item[] Question: Does the research conducted in the paper conform, in every respect, with the NeurIPS Code of Ethics \url{https://neurips.cc/public/EthicsGuidelines}?
    \item[] Answer: \answerYes{} 
    \item[] Justification: The research conducted in the paper conforms with the NeurIPS Code of Ethics.

\item {\bf Broader impacts}
    \item[] Question: Does the paper discuss both potential positive societal impacts and negative societal impacts of the work performed?
    \item[] Answer: \answerYes{} 
    \item[] Justification: paper discuss potential positive societal impacts of the work performed.
    
\item {\bf Safeguards}
    \item[] Question: Does the paper describe safeguards that have been put in place for responsible release of data or models that have a high risk for misuse (e.g., pretrained language models, image generators, or scraped datasets)?
    \item[] Answer: \answerNA{} 
    \item[] Justification: The paper poses no such risks

\item {\bf Licenses for existing assets}
    \item[] Question: Are the creators or original owners of assets (e.g., code, data, models), used in the paper, properly credited and are the license and terms of use explicitly mentioned and properly respected?
    \item[] Answer: \answerYes{} 
    \item[] Justification: Assets are the license and terms of use explicitly mentioned and properly respected.

\item {\bf New assets}
    \item[] Question: Are new assets introduced in the paper well documented and is the documentation provided alongside the assets?
    \item[] Answer: \answerNA{} 
    \item[] Justification:  The paper does not release new assets.

\item {\bf Crowdsourcing and research with human subjects}
    \item[] Question: For crowdsourcing experiments and research with human subjects, does the paper include the full text of instructions given to participants and screenshots, if applicable, as well as details about compensation (if any)? 
    \item[] Answer: \answerNA{} 
    \item[] Justification: The paper does not involve crowdsourcing nor research with human subjects.

\item {\bf Institutional review board (IRB) approvals or equivalent for research with human subjects}
    \item[] Question: Does the paper describe potential risks incurred by study participants, whether such risks were disclosed to the subjects, and whether Institutional Review Board (IRB) approvals (or an equivalent approval/review based on the requirements of your country or institution) were obtained?
    \item[] Answer: \answerNA{} 
    \item[] Justification:The paper does not involve crowdsourcing nor research with human subjects.

\item {\bf Declaration of LLM usage}
    \item[] Question: Does the paper describe the usage of LLMs if it is an important, original, or non-standard component of the core methods in this research? Note that if the LLM is used only for writing, editing, or formatting purposes and does not impact the core methodology, scientific rigorousness, or originality of the research, declaration is not required.
    \item[] Answer: \answerNA{} 
    \item[] Justification: The core method development in this research does not involve LLMs as any important, original, or non-standard components.

\end{enumerate}

\end{document}